\documentclass{article}

\PassOptionsToPackage{numbers}{natbib}
\usepackage[preprint]{neurips_2026}

\usepackage[utf8]{inputenc} 
\usepackage[T1]{fontenc}    
\usepackage{hyperref}       
\usepackage{url}            
\usepackage{booktabs}       
\usepackage{amsfonts}       
\usepackage{nicefrac}       
\usepackage{microtype}      
\usepackage{xcolor}         
\usepackage[acronym]{glossaries} 
\usepackage{multirow} 
\usepackage{graphicx}

\title{Geometry Without Coordinates: LiDAR Diffusion as a 3D Feature Bridge}

\author{%
  Samed Do\u{g}an\\
  Department of Electrical Engineering and Information Technology\\
  Munich University of Applied Sciences\\
  Munich, Bavaria 80335 \\
  \texttt{samed.dogan@hm.edu} \\
  \And
  Nico Leuze\\
  Department of Electrical Engineering and Information Technology\\
  Munich University of Applied Sciences\\
  Munich, Bavaria 80335 \\
  \texttt{nico.leuze@hm.edu}
  \And
  Alfred Schöttl\\
  Department of Electrical Engineering and Information Technology\\
  Munich University of Applied Sciences\\
  Munich, Bavaria 80335 \\
  \texttt{alfred.schoettl@hm.edu}\\
}

\newacronym{ddpm}{DDPM}{Denoising Diffusion Probabilistic Model}
\newacronym{miou}{MIoU}{Mean Intersection over Union}
\newacronym{iou}{IoU}{Intersection over Union}
\newacronym{absrel}{AbsRel}{Absolute Mean Relative Error}
\newacronym{rmse}{RMSE}{Root Mean Square Error}
\newacronym{pca}{PCA}{Principal Component Analysis}

\begin{document}

\maketitle

\begin{abstract}
Transferring the rich priors of large 2D foundation models to sparse 3D LiDAR remains challenging, as training native 3D foundation models at comparable scale is limited by data and annotation scarcity. We introduce a LiDAR-conditioned diffusion model trained on pseudo-labels from off-the-shelf 2D foundation models. The model supports multiple output modalities, including depth, semantic segmentation and instance prediction, selectable via a textual task prompt. Because the model is conditioned on LiDAR, both its outputs and its intermediate UNet features can be projected back onto the input point cloud, enabling analysis of a 3D representation learned entirely under 2D supervision.
We study this representation directly in point-cloud space, explicitly excluding raw spatial coordinates to isolate feature content from projection geometry. Linear probes recover up to $\sim$23\% \gls{miou} on 3D semantic classes, compared to $\sim$3.5\% for a matched Gaussian-noise control, indicating substantial non-trivial structure. Pairwise cosine similarity across modality-specific feature streams reveals a layered organization. Early encoder layers remain weakly aligned across modalities while individually decodable, intermediate layers converge toward a shared representation, and decoder layers re-specialize toward task-specific outputs.
These findings indicate that LiDAR-conditioned diffusion models can induce structured 3D representations from 2D supervision alone, with a modality-dependent manifold that locally unifies near a shared bottleneck. This positions diffusion as a viable mechanism for transferring large-scale 2D priors into sparse 3D domains.
\end{abstract}

\section{Introduction}
\label{sec:intro}
Acquiring large-scale, well-aligned multi-modal 3D data remains a central bottleneck for LiDAR perception. Established benchmarks such as nuScenes\cite{nuscenes}, Waymo Open Dataset\cite{waymo}, and KITTI\cite{kitti-raw} provide synchronized camera-LiDAR data, but at a scale far below what is available for 2D vision. These constraints have motivated a growing body of work on generative modeling for sensor data, particularly diffusion-based frameworks inspired by \cite{diffusion1-ddpm, diffusion2-ldm, diffusion3-earlywork}. Recent generative approaches model
camera images \cite{camera1-magicdrive, camera2-magicdrive2, camera3-bevcontrol, camera4-drivedreamer, camera5-vista, camera6-panacea}, LiDAR point clouds \cite{lidar1-lidarDM, lidar2-r2dm, lidar3-lidargen, lidar4-upsamling, lidar5-LiDM, lidar6-rangeldm, lidar7-text2lidar, lidar8-lidarcrafter}, and joint multi-modal distributions \cite{lc1-xdrive, lc2-holodrive, lc3-uniscene, lc4-bevworld} primarily as sensor simulators, aiming to reproduce raw observations rather than transferable representations.\\
In contrast, the 2D image domain has benefited from large-scale pretraining on foundation models such as DINO\cite{foundation11-dino, foundation1-dinov2, foundation2-dinov3}, Depth Anything\cite{foundation3-depthanything3, foundation4-depthanything2}, and Segment Anything\cite{foundation5-sam2, foundation6-sam3}. These models encode rich priors over geometry, semantics, and object boundaries learned from web-scale image corpora. Comparable priors are largely absent in 3D, where early efforts exist \cite{zhou2023uni3d, wu2025sonata, zhang2026utonia, liu2023openshape}, but remain orders of magnitude smaller than their 2D counterparts. 3D data is sparse, irregular, and significantly more costly to scale. This disparity raises a central question: can the priors learned by 2D foundation models be transferred into 3D representations without training a native 3D model from scratch?
We approach this problem with a diffusion model that bridges the two domains by conditioning on LiDAR. Rather than treating diffusion as a sensor simulator, we use it as a vehicle for transporting 2D priors into 3D: the network learns to produce dense scene representations conditioned on sparse 3D measurements, supervised entirely by pseudo-labels from
off-the-shelf 2D foundation models. A single backbone produces multiple modalities, namely depth, semantic segmentation, and instance prediction, switched by textual prompt. The LiDAR conditioning spatially aligns dense 2D priors with 3D observations by construction, which makes intermediate
features directly reprojectable into the input point cloud.\\
We then ask: what structure does this transfer induce? To answer this, we probe intermediate features directly in point-cloud space, explicitly excluding raw spatial coordinates to isolate feature content from projection geometry. Our analysis reveals two findings. First, the learned
representation encodes meaningful 3D structure: linear probes reach $\sim$23\% \gls{miou} on 3D semantic classes, compared to $\sim$3.5\% for a Gaussian-noise control. Second, modality-conditioned features exhibit a consistent organization across network depth: early layers remain
modality-specific, intermediate layers converge toward a shared representation, and later layers re-specialize toward task-specific outputs. Rather than collapsing modalities, the network arranges them along its depth.\\
We summarize our contributions as follows:
\begin{itemize}
\item We introduce a LiDAR-conditioned diffusion framework that produces
dense scene representations across multiple task modalities through
textual prompting, enabling pseudo-label-based transfer of 2D
foundation-model priors into 3D point clouds.
\item We propose a probing protocol that evaluates diffusion features
directly in point-cloud space while excluding raw spatial coordinates,
isolating what the features encode from what the projection geometry leaks.
\item We show that, despite the absence of 3D ground-truth labels, the
learned features encode structured 3D information and are organized across
network depth into modality-specific streams that converge near a shared
bottleneck.
\end{itemize}
\section{Related Work}
\subsection{Diffusion Models for Conditional and Structured Generation}
\glspl{ddpm} introduced by \cite{diffusion1-ddpm} have become a dominant paradigm for high-fidelity generative modeling. Subsequent works extended diffusion models to conditional generation, including text conditioning \cite{diffusion8-t2i-gligen, diffusion9-t2i-glide, diffusion10-t2i-imagen}, and spatial guidance \cite{diffusion17-spatial1, diffusion18-spatial2, diffusion19-spatial3}. Text-to-image models such as Stable Diffusion \cite{diffusion2-ldm} and DALL.E 2 \cite{diffusion11-dalle} demonstrate the scalability of diffusion under large-scale supervision.
 Adapters such as ControlNet \cite{diffusion5-controlnet}, T2I \cite{diffusion6-t2i} and IP-Adapter \cite{diffusion7-ipadapter} introduce spatial conditioning mechanism, while classifier-free guidance \cite{diffusion12-cfg} enables flexible conditioning without auxiliary networks. Beyond image synthesis, diffusion has also been applied to depth estimation \cite{diffusion13-depth}, segmentation \cite{diffusion14-segmentation}, and other dense prediction tasks \cite{diffusion15-flow, diffusion16-light}, blurring the boundary between generative modeling and discriminative structured prediction.
\subsection{Diffusion Features for Perception}
A growing line of work shows that the intermediate U-Net features of pretrained text-to-image diffusion models encode strong perceptual priors that transfer to discriminative tasks. \cite{diff-features1} adapt Stable Diffusion\cite{diffusion2-ldm} features for dense visual perception with task-specific heads, while~\cite{diff-features2} use them for open-vocabulary panoptic
segmentation by pairing them with text encoders. Subsequent work treats the features themselves as a primary object of study: \cite{diff-features5} aggregate features across timesteps and decoder
levels to identify the layers where semantic structure concentrates, \cite{diff-features3} demonstrate that diffusion features support zero-shot semantic correspondence, and~\cite{diff-features4} address the mismatch between training-time noised features and inference-time clean
inputs. Closer to the dense-prediction setting,~\cite{diff-features6} use diffusion features for semantic segmentation without task-specific finetuning.\\
These works establish that diffusion features carry transferable
perceptual structure, but they study a single output modality at a time
on 2D images. Two questions are therefore open. First, how do these
features behave when the same backbone is conditioned to produce multiple
output modalities at inference? Second, do the same observations transfer
when the input modality is sparse 3D rather than dense RGB? Our work
addresses both: we analyze a single LiDAR-conditioned diffusion model
that produces depth, semantic, and instance outputs through textual
prompting, and we probe its features in 3D point-cloud space rather than
2D pixel space. The cross-modal organization we observe (Section~\ref{sec:cosine})
is not visible from any single-modality probing setup.
\section{Preliminaries}
\paragraph{Latent Diffusion Models.}
We use Stable Diffusion 1.5~\cite{diffusion2-ldm}, a latent diffusion model that performs the forward and reverse diffusion process in the latent space of a pretrained autoencoder $(\mathcal{E}, D)$. The denoising network $\epsilon_\theta(z_t, t, c)$ is trained to predict
noise added to latents $z_0 = \mathcal{E}(x_0)$ under conditioning $c$. In our setting, $c$ comprises the projected LiDAR view (depth and intensity) and a textual task prompt that selects the target representation type. Full derivations are provided in Appendix~\ref{app:diffusion}.
\paragraph{LiDAR-Camera Projection.}
We assume calibrated LiDAR-camera pairs with known intrinsic $K$ and extrinsic $T_{CL}$. Throughout the paper, we denote the composite projection of a 3D LiDAR point $\mathbf{p}_L$ to pixel coordinates as $\Pi(T_{CL}, \mathbf{p}_L) \in \mathbb{R}^2$, and its inverse, given
a pixel $\mathbf{u}$ and a depth value $\lambda$, as $\Pi^{-1}(T_{CL}, \mathbf{u}, \lambda) \in \mathbb{R}^3$. Full projection and backprojection equations are provided in Appendix~\ref{app:projection}.
\section{Methodology}
\subsection{Pseudo-label Generation}
Let $I\in\mathbb{R}^{H\times W \times3}$ denote a camera image and $\mathcal{P}=\{\mathbf{p}_i\}_{i=1}^N$ a corresponding point set in the LiDAR coordinate frame.
We obtain supervision signals by applying pretrained 2D foundation models $\{\mathcal{F}_k\}$ to each image $y_k=\mathcal{F}_k(I)$, where $y_k$ may represent predictions of
varying modality such as dense depth, semantic segmentation, or instance
masks. Specifically, we use Depth Anything~v3~\cite{foundation3-depthanything3} for dense monocular depth, Segment Anything~2~\cite{foundation5-sam2} for instance masks, and a SegFormer~\cite{xie2021segformer} model finetuned on Cityscapes~\cite{cityscapes} for semantic segmentation. The predictions are generated offline and remain fixed during diffusion training, acting as supervision targets. Our structured multi-modal representation takes on the form of $x_0=\Phi(\{y_k\})$, where $\Phi$ denotes concatenation across the batch-axis. The diffusion model is trained to model the conditional distribution
\begin{equation}
    p_{\theta}(x_0|\Pi(T_{LC}, \mathcal{P}), s)
\end{equation}
with $\Pi(T_{LC}, \mathcal{P})$ representing the projection of the point cloud depth and intensity onto the corresponding camera view of $I$ and $s$ the desired representation type.\\
Foundation-model predictions may contain noise or inconsistencies with ground-truth sensor measurements. We apply modality specific preprocessing to ensure geometric and semantic consistency, using metadata available within the nuScenes\cite{nuscenes} dataset. We also accommodate for the input dimensionality of the encoder $\mathcal{E}$, which expects a 3-channel image.
\subsubsection{Depth}
We learn relative depth and recover metric scale post-hoc via least-squares alignment to LiDAR (Section~\ref{sec:depth}). For each pseudo-depth map $d$,  sky pixels~\cite{foundation3-depthanything3} are reassigned to the $0.99$-quantile $d_{0.99}$, yielding $d'$. We then apply a logarithmic  transform and quantile-based normalization~\cite{diffusion13-depth}:
\begin{equation}
    \tilde{d} = 2 \cdot\text{clip}\!\left(\frac{\log(d')-d_{\min}}{d_{\max}-d_{\min}}, 0, 1\right)-1,
\end{equation}
with $d_{\min}=\log(d')_{0.02}$, $d_{\max}=\log(d')_{0.98}$. The result is replicated to 3 channels to match $\mathcal{E}$.
\subsubsection{Instance Segmentation}
\label{sec:instance-seg}
To initialize the instance segmentation, we leverage existing bounding box annotations of road users as spatial anchors, iteratively refining the segmentation masks based on prior predictions. During training, we encode the discrete instance masks into a a continuous, dense representation suitable for diffusion. Specifically, we adapt the the center-regression approach from \cite{panoptic-deeplab} to the diffusion domain. The segmentation mask is parameterized as a three-channel tensor: a foreground mask $M\in\{0, 1\}$ and two relative offset vectors $O_x\in[-1,1], O_y\in[-1, 1]$. The offset vectors are normalized by the spatial image dimension and point to the center of mass of its respective instance.\\
At inference, we decode the per-view continuous predictions into a single set of point-cloud instance IDs. For each LiDAR point that projects into a camera view, the predicted offset vector is added to its pixel coordinate to obtain the predicted centre location in pixel space, which is then unprojected into the LiDAR coordinate frame assuming the centre lies at the same depth as the surface point. Predicted centres and foreground confidences are mean-fused across the views in which a point is visible. Points whose fused foreground confidence exceeds a threshold $\tau_{\text{fg}} = 0.8$ are retained as instance candidates, and their predicted centres are clustered in 3D LiDAR space using DBSCAN~\cite{dbscan} with a Euclidean neighbourhood radius
of $\epsilon = 1\,\text{m}$ and minimum cluster size of $5$ points. Each resulting cluster yields one instance ID.
\subsubsection{Semantic Segmentation}
We convert the class labels to fixed \textit{RGB} colors using the conversion table provided by the Cityscapes \cite{cityscapes} dataset. For inference, we retrieve the class indexes via nearest-neighbor color quantization, using the conversion table as color palette.
\subsection{Conditioning Dropout Ablation}
\label{sec:dropout}
The conditioning input has two LiDAR channels: depth $D$ and reflective 
intensity $R$. We compare five training regimes to characterize the network's 
reliance on each channel: (1) \textit{no dropout} (baseline), (2)
\textit{depth-channel dropout}, (3) \textit{intensity-channel dropout}, 
(4) \textit{random per-channel dropout} (one of the two channels, never both), 
and (5) \textit{joint pixel-space dropout} (per-pixel mask applied to both 
channels simultaneously). All channel-level regimes use $p=0.20$; (5) uses 
per-pixel $p=0.20$. Regimes (2-4) corrupt entire channels at the channel 
level; (5) corrupts both channels at the pixel level, so neither modality is 
ever fully present nor fully absent. These four regimes serve as ablation 
conditions for Section~\ref{sec:feature_restructuring}.
\subsection{Probing Methodology}
\label{sec:probing}
We assess the quality of internal U-Net representations by evaluating them in their final downstream domain: the 3D LiDAR point cloud. Our evaluation pipeline proceeds as follows: (i) we generate dense 2D views from sparse LiDAR inputs, (ii) extract feature maps from all U-Net
levels, and (iii) reproject these features back onto the original 3D point cloud via grid sampling. We then train a linear probe on the reprojected point-wise features.
The probe is a single linear layer mapping per-point feature vectors to per-class logits, trained for $60$ optimization steps. At each step, features and pseudo-labels are accumulated across a fixed training chunk of $10$ samples, and the gradient is computed over the concatenated set; we use full-batch updates rather than stochastic minibatches. Evaluation is performed on a disjoint $10$-sample bucket sampled from the unobserved validation set. We report mean intersection-over-union over the nuScene lidarseg dataset. To isolate the semantic and geometric information encoded exclusively within the diffusion features, we exclude raw spatial coordinates $(x, y, z)$ from the probe's input. As a comparative baseline, we construct a Gaussian-noise control feature: per-channel mean and standard deviation are computed from the U-Net features at each level, and a tensor of matched shape is sampled from $\mathcal{N}(\mu_c, \sigma_c)$ as the probe input. Because the control preserves the marginal statistics of the real features but contains no spatial structure, near-chance probe performance on this control verifies that probe scores reflect the structure of the U-Net features rather than incidental properties of the projection geometry.
\section{Experiments}
\subsection{Diffusion Features Encode 3D Relevant Information}
\label{sec:long-probe}
\begin{figure}
    \centering
    \includegraphics[width=0.8\linewidth]{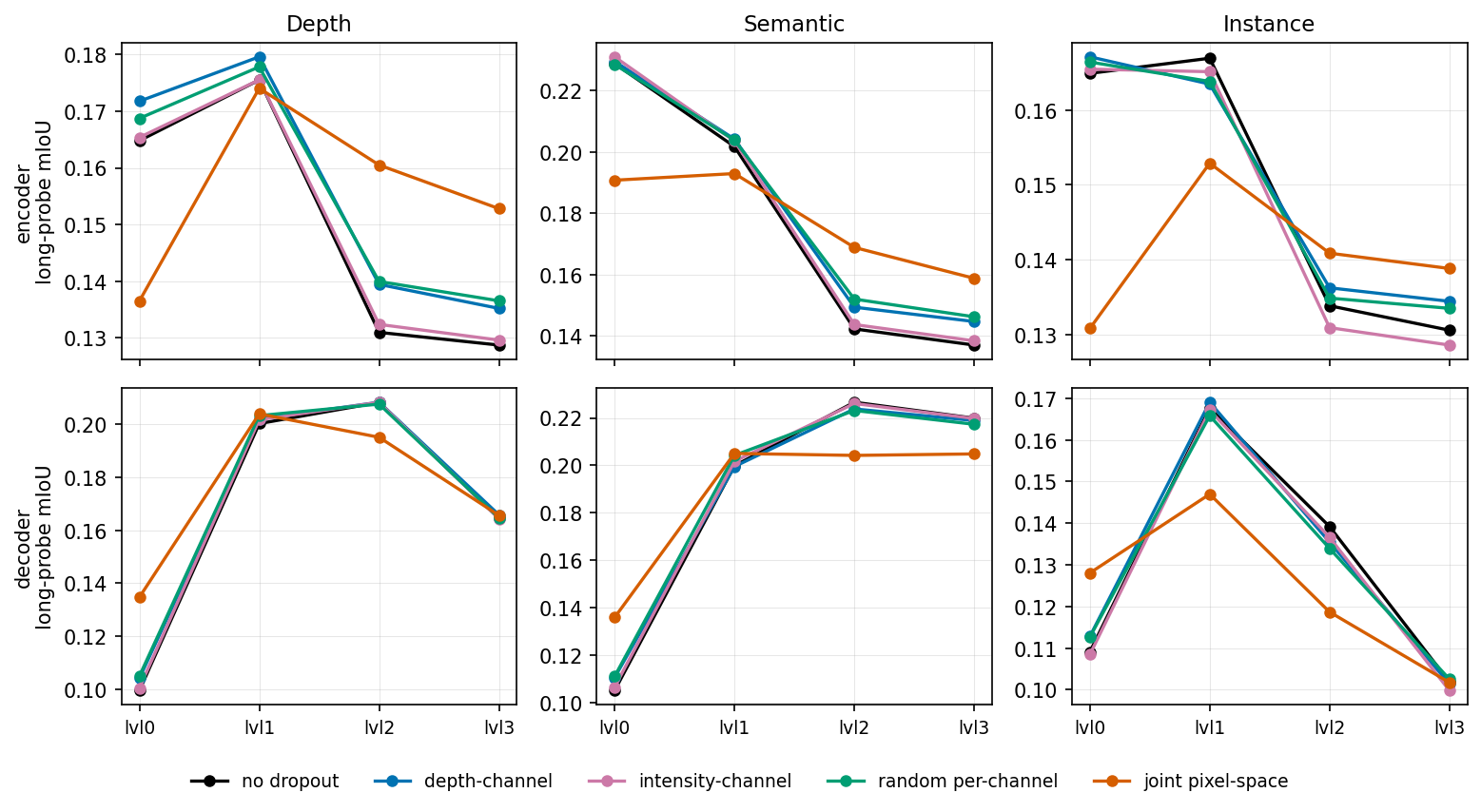}
    \caption{Long-probe \gls{miou} per U-Net level, separately for encoder and decoder features. Encoder probes peak at shallow levels (L0--L1) and decay monotonically with depth. Decoder probes show an inverted-U shape, peaking at L1--L2. Joint pixel-space dropout is the only ablation that visibly shifts these patterns, suppressing shallow encoder levels and collapsing all best-level probes onto L1}
    \label{fig:per-level-probe}
\end{figure}

To dissect the impact of our representational modeling, we evaluate the long-range generalization of the learned U-Net features across all U-Net levels. Table~\ref{tab:headline_long_probes} reports long-range probing \gls{miou} on the unseen validation set. Each cell reports the score at the level where the probe is strongest, selected independently per ablation, stage, and feature type. This isolates the question of \textit{whether} features encode task-relevant 3D structure. The complementary question of \textit{where} in the network this information lives is shown in Figure~\ref{fig:per-level-probe}, which plots probe \gls{miou} as a function of U-Net level for both stages. We additionally probed level-concatenated features, stacked along the channel dimension. These concatenated probes consistently underperformed the best single level probe. We attribute this to the linear probe's limited capacity to fit a $4\times$ higher-dimensional input under the same regularization, rather than to a property of the underlying representation.
\begin{table}
\centering
\small
\caption{Long linear-probe mIoU on backprojected UNet features, evaluated in
point-cloud space without xyz coordinates. Reported at the strongest single
level (subscript), selected independently per ablation, stage and feature type from
Figure~\ref{fig:per-level-probe}. The \textsc{Base} column is a
Gaussian-noise control feature; its near-chance score confirms probes do
not exploit projection geometry.}
\label{tab:headline_long_probes}
\newcommand{\val}[2]{$#1_{\,\textsc{\scriptsize #2}}$}
\begin{tabular}{l cccc cccc}
\toprule
& \multicolumn{4}{c}{Encoder features} & \multicolumn{4}{c}{Decoder features} \\
\cmidrule(lr){2-5}\cmidrule(lr){6-9}
Dropout & Depth & Sem. & Inst. & Base & Depth & Sem. & Inst. & Base \\
\midrule
None       & \val{0.176}{l1} & \val{0.229}{l0} & \val{0.167}{l1} & \val{0.035}{--} & \val{0.208}{l2} & \val{0.227}{l2} & \val{0.168}{l1} & \val{0.038}{--} \\
Depth      & \val{0.180}{l1} & \val{0.230}{l0} & \val{0.167}{l0} & \val{0.036}{--} & \val{0.208}{l2} & \val{0.224}{l2} & \val{0.169}{l1} & \val{0.037}{--} \\
Intensity  & \val{0.176}{l1} & \val{0.231}{l0} & \val{0.166}{l0} & \val{0.035}{--} & \val{0.208}{l2} & \val{0.226}{l2} & \val{0.167}{l1} & \val{0.036}{--} \\
Random & \val{0.178}{l1} & \val{0.229}{l0} & \val{0.166}{l0} & \val{0.036}{--} & \val{0.208}{l2} & \val{0.223}{l2} & \val{0.166}{l1} & \val{0.035}{--} \\
Pixel Space  & \val{0.174}{l1} & \val{0.193}{l1} & \val{0.153}{l1} & \val{0.034}{--} & \val{0.204}{l1} & \val{0.205}{l1} & \val{0.147}{l1} & \val{0.034}{--} \\
\bottomrule
\end{tabular}
\end{table}
Across all conditioning regimes, linear probes recover task-relevant 3D structure from UNet features. The strongest probe reaches $0.231$ \gls{miou} on semantic segmentation, substantially above the Gaussian-noise control, which remains at $0.035-0.038$ across every ablation and every task. This $\sim6\times$ gap establishes that the probes are reading task-relevant content from the features themselves rather than exploiting projection geometry. 
Figure~\ref{fig:per-level-probe} reveals that encoder and decoder features organize this information differently along the network. Encoder probes decay monotonically with depth across all three feature types: semantic \gls{miou} drops from $0.229$ at level $0$ to $0.137$ at level$3$, a $40\%$ relative loss. Decoder probes invert this, with shallow features (L0) at near-chance levels ($0.10$--$0.11$ \gls{miou} and a clear peak at level$1$--level$2$. This asymmetry is consistent with the U-Net's architectural role: the encoder compresses input-side geometry into progressively abstract features, while the decoder reconstruct spatial structure most decodably at mid-resolution. We examine the relationship between modality streams within each level in Section~\ref{sec:cosine}. Joint pixel-space dropout (orange in Figure~\ref{fig:per-level-probe}) is the only regime that visibly shifts this pattern, suppressing the shallow-encoder peak and collapsing all of its best-level probes onto level$1$. We examine this in Section~\ref{sec:feature_restructuring}.
\subsection{Modality Stream Analysis}
\label{sec:cosine}
\begin{figure}
    \centering
    \includegraphics[width=0.8\linewidth]{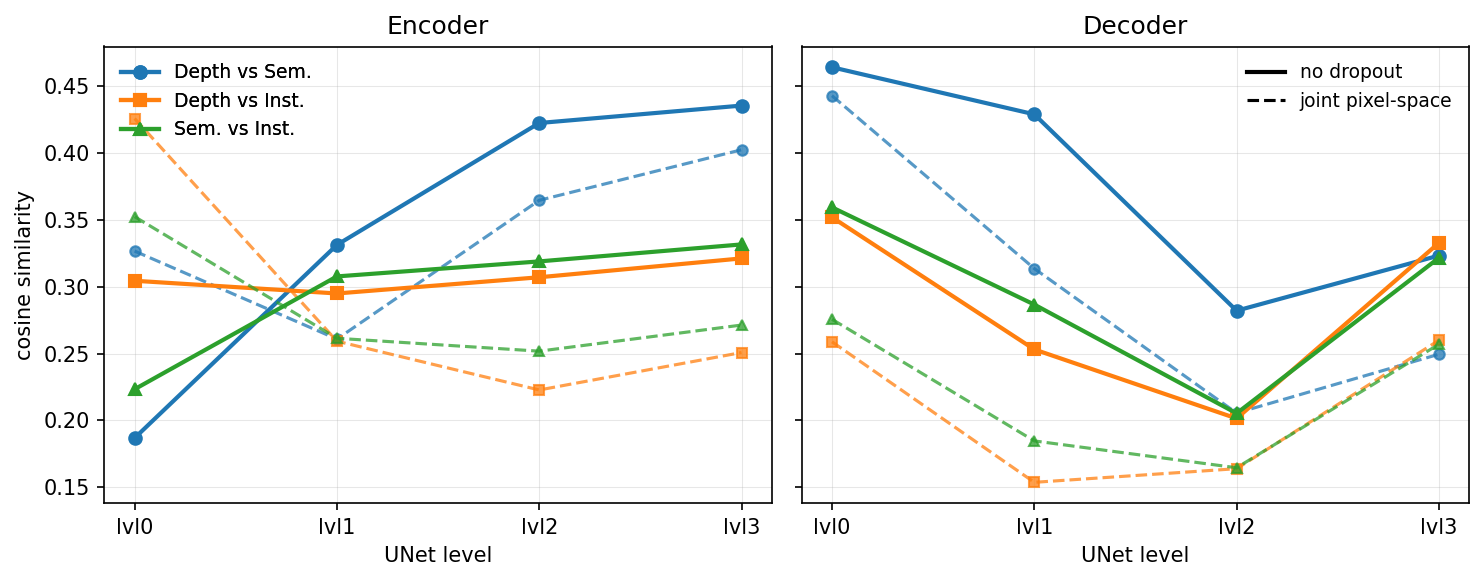}
    \caption{Pairwise cosine similarity of features (baseline) compared to features streams produced by pixel-wise channel dropout. Joint pixel-space dropout makes shallow encoder representations more unified, while deeper layers become more specialized.}
    \label{fig:cosine_base_vs_both}
\end{figure}

Section~\ref{sec:long-probe} established that U-Net features at every level encode task-relevant 3D structure, and that encoder and decoder stages reach peak decodability at different depths. Linear probing, however, treats each modality stream in isolation: it measures whether information is present, not how the streams produced under different task prompts relate to one another. A single backbone that switches modality through text prompt could in principle produce nearly identical features regardless of prompt, collapsing into a single representation that the prompt merely relabels, or it could maintain three substantially distinct streams. To distinguish these cases, we measure pairwise cosine similarity between feature maps produced under each pair of task prompts, computed at every U-Net level on the no-dropout model. Low similarity indicates that the streams occupy different directions in feature space. High similarity indicates convergence towards a shared representation. We report cosine-similarity for the baseline in Table~\ref{tab:cosine_main}, as we observed no substantial feature restructuring across the ablations, except for the pixel level joint-dropout. We further provide a direct comparison between those two ablations in Figure~\ref{fig:cosine_base_vs_both} and discuss the direct influence on feature structure more thoroughly in Section~\ref{sec:feature_restructuring}.
\begin{table}
\centering
\caption{Pairwise cosine similarity between modality-conditioned features
(no-dropout model). Encoder similarity grows with depth; decoder similarity
drops to a minimum at level 2 with a small rebound at level 3.}
\label{tab:cosine_main}
\begin{tabular}{l cccc cccc}
\toprule
& \multicolumn{4}{c}{Encoder} & \multicolumn{4}{c}{Decoder} \\
\cmidrule(lr){2-5}\cmidrule(lr){6-9}
Pair & L0 & L1 & L2 & L3 & L0 & L1 & L2 & L3 \\
\midrule
Depth vs.\ Semantic   & 0.187 & 0.331 & 0.422 & 0.435 & 0.464 & 0.429 & 0.282 & 0.323 \\
Depth vs.\ Instance   & 0.304 & 0.295 & 0.307 & 0.321 & 0.352 & 0.253 & 0.202 & 0.333 \\
Semantic vs.\ Instance & 0.223 & 0.308 & 0.319 & 0.332 & 0.359 & 0.287 & 0.205 & 0.322 \\
\bottomrule
\end{tabular}
\end{table}
Figure~\ref{fig:cosine_base_vs_both} and Table~\ref{tab:cosine_main} show two opposing trends. In the encoder, cross-modal similarity grows with depth: depth-vs-semantic alignment rises from $\sim0.19$ at level 0 to $\sim0.44$ at level 3, and the other two pairs follow the same monotonic pattern. Shallow encoder layers maintain the most modality-decomposed representations of any point in the network, while progressively deeper layers converge toward a shared manifold. The decoder reverses this trend: similarity is highest at level 0 $\sim$($0.36$--$0.46$) across the three pairs, drops to a minimum at level 2 $\sim$($0.2$--$0.28$), and shows small rebound at level 3 $\sim$($0.32$--$0.33$). This pattern is consistent with the encoder progressively unifying modality-specific inputs into a shared bottleneck representation, and the decoder re-specializing those features as it reconstructs modality-specific outputs. Since the weights produce all three streams, our observation is a structure that the architecture allows, but not enforces. Together with the per-level decodability results of Section~\ref{sec:long-probe}, these measurements show that the U-Net does not collapse modalities into a single representation. It organizes them into a modality-decomposed manifold that locally unifies near the bottleneck.
\subsection{Feature Restructuring} \label{sec:feature_restructuring}
\begin{figure}
    \centering
    \includegraphics[width=0.8\linewidth]{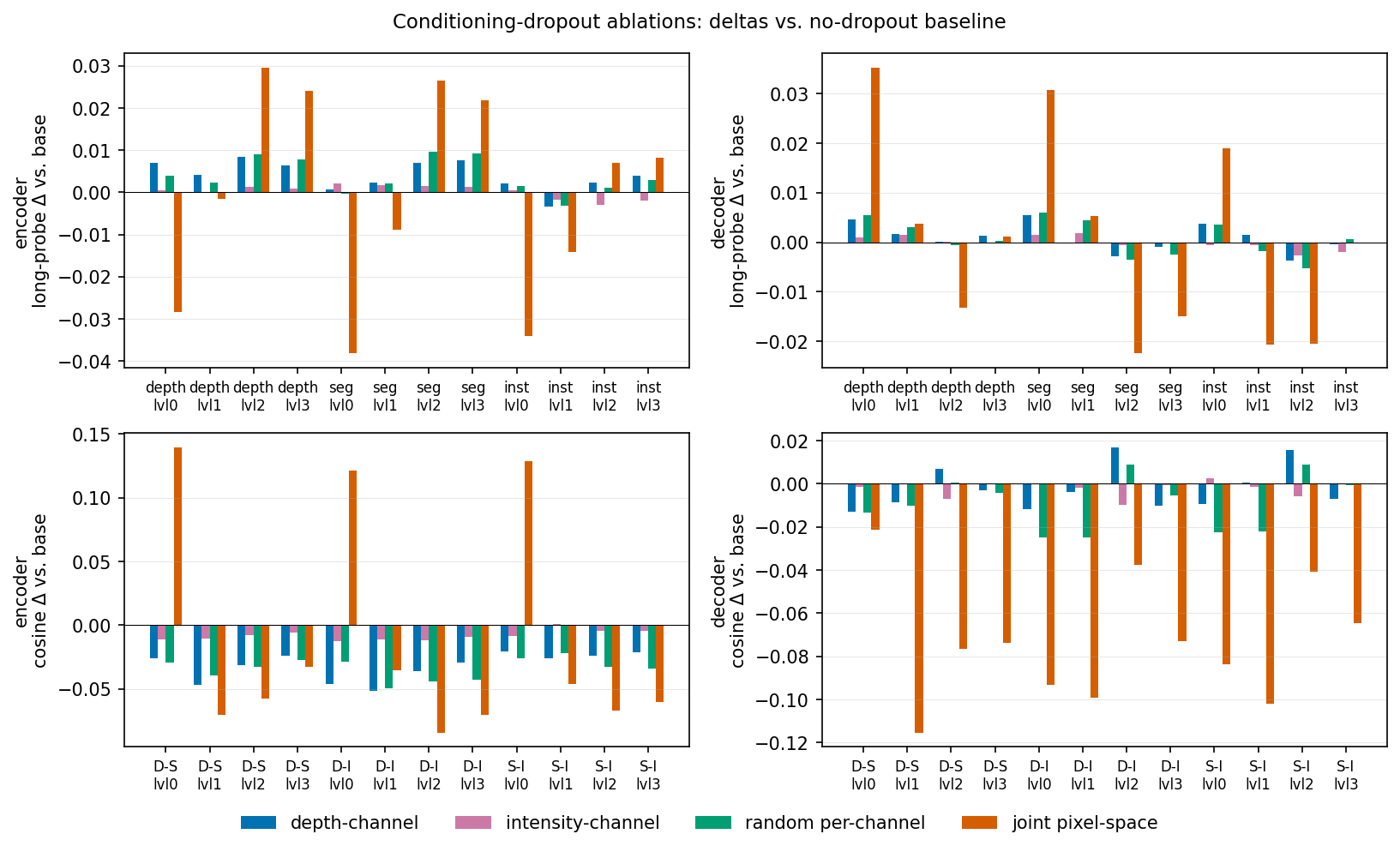}
    \caption{Conditioning-dropout effects, reported as deltas relative to the no-dropout baseline. Top row: long-probe \gls{miou} differences per feature, type and level. Bottom row: pairwise cosine-similarity differences per modality, pair and level. Single-channel dropout cluster near zero across both metrics. Joint pixel-space dropout is the only regime that consistently displaces the representation. It suppresses shallow-encoder probe quality and raises shallow-encoder cosine alignment, while reducing decoder cosine alignment across every level and pair. D-S, D-I and S-I denote Depth--Semantic, Depth--Instance, and Semantic--Instance pairs.}
    \label{fig:deltas}
\end{figure}

The preceding two sections used the no-dropout model to establish where information lives in the U-Net (Section~\ref{sec:long-probe}) and how modality streams are organized along its depth (Section~\ref{sec:cosine}). We now ask how the conditioning regime shapes that organization. Specifically, we ask what part of the LiDAR input the network must learn to be robust. Recall that we condition on a two-channel projected LiDAR view: sparse-depth and reflective intensity. We compare four dropout regimes against the no-dropout baseline: \textit{depth} drops the depth channel entirely with some probability, \textit{intensity} drops intensity entirely, \textit{random} drops one of the two channels at random, and \textit{joint pixel-space} masks both channels at the pixel level. Figure~\ref{fig:deltas} reports each ablation as a delta against \textit{base}, separately for long-probe \gls{miou} and pairwise cosine similarity, at every level and stage.
Three of the four ablations cluster tightly near zero across every panel: depth-channel, intensity-channel, and random per-channel dropout produce probe-\gls{miou} shifts within $\pm0.01$ and cosine-similarity shifts within $\pm0.07$, with no consistent direction across levels or pairs. The intensity-channel result in particular is essentially a no-op, indicating that the network treats the depth channel as the dominant 3D conditioning signal. Joint pixel-space dropout, in contrast, produces a coordinated restructuring. In the encoder it suppresses shallow probe quality while raising probe quality at deeper levels. The corresponding cosine deltas are large and level 0 concentrated. In the decoder, all three cosine pairs drop substantially across every level. We hypothesize that the absence of reliable conditioning information in either channel forces shallow encoding layers to learn more globally-mixed denoising representations. Task-discriminative content is deferred to deeper layers, and the decoder becomes more modality-decomposed throughout. We interpret this as the network trading shallow modality specialization for input-noise robustness, with the cost being a small but measurable drop in downstream output quality (Table~\ref{tab:direct_metrics}).
\subsection{Output Quality}
We evaluate outputs only insofar as they reflect properties of the learned representations. To this end, we report quantitative depth metrics, which provide a direct and minimally confounded probe of geometric structure. These metrics confirm that the representational changes induced by joint-pixel-space dropout (Section~\ref{sec:feature_restructuring}) translate consistently to the output space. For semantic and instance predictions, we focus on qualitative evaluation. Quantitative evaluation in these settings depends on factors orthogonal to our study. Namely, cross-dataset label alignment (Cityscapes $\rightarrow$ nuScenes lidarseg) and post-processing design for instance extraction, introduce substantial variance unrelated to representation quality. We therefore restrict quantitative analysis to depth, where measurement more directly reflects the underlying features.
\subsubsection{Depth} \label{sec:depth}
We asses model performance in dense depth estimation by adopting the \gls{absrel} and $\delta1$ metric as well-known evaluation protocols observed in \cite{depth-least-squares, depth1-eval, depth2-eval, depth3-eval, depth4-eval, foundation4-depthanything2}. We compute \gls{absrel} as absolute relative distance over all observed LiDAR points $M$: $\frac{1}{M}\sum_{i=1}^M|\hat{d}_i-d_i|/d_i$ and $\delta1$ accuracy as the ratio of all points satisfying $\max(\hat{d}_i/d_i, d_i/\hat{d}_i) < 1.25$. Additionally, we report the \gls{rmse} $\frac{1}{\sqrt{M}}\|d-\hat{d}\|$ to reflect prediction accuracy in far-field regions, where observed data is typically very sparse and noisy. All tests are conducted in metric LiDAR space using least-squares alignment following \cite{diffusion13-depth, depth-least-squares}. Concretely, we use LiDAR depth $d$ as sparse measurements and solve for the affine transformation $\hat{d} = \tilde{d} \times s + t$, where $\tilde{d}$ is the predicted affine invariant depth, $\hat{d}$ the aligned metric depth and $s, t$ depict scale and shift transforms, respectively.
\begin{table}
\centering
\small
\caption{Depth metrics computed on LiDAR points after least-squares
metric alignment}
\label{tab:direct_metrics}
\begin{tabular}{l ccc}
\toprule
Dropout & abs\_rel $\downarrow$ & RMSE $\downarrow$ & $\delta_{1.25} \uparrow$ \\
\midrule
none       & \textbf{0.160} & \textbf{7.11} & \textbf{0.781} \\
depth      & 0.162 & 7.17 & 0.778 \\
intensity  & 0.161 & 7.11 & 0.780  \\
random per-channel & 0.162 & 7.17 & 0.778  \\
joint pixel-space  & 0.165 & 7.27 & 0.777  \\
\bottomrule
\end{tabular}
\end{table}
\subsubsection{Qualitative Results}
\begin{figure}
\centering
\setlength{\tabcolsep}{2pt}
\begin{tabular}{@{}ccc@{}}
\small Depth Prompt & \small Semantic Prompt & \small Instance Prompt \\[2pt]
\includegraphics[width=0.31\linewidth]{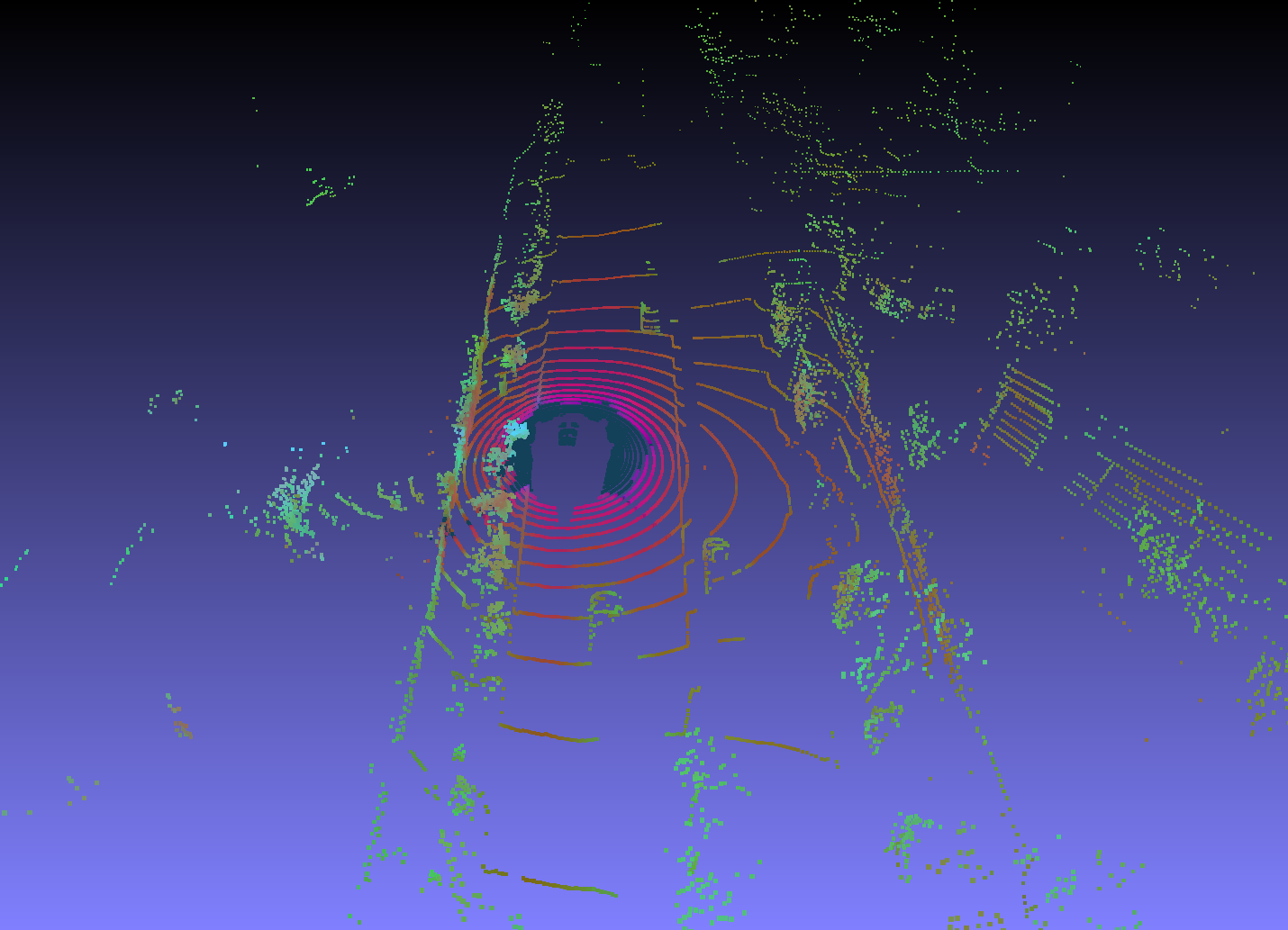} &
\includegraphics[width=0.31\linewidth]{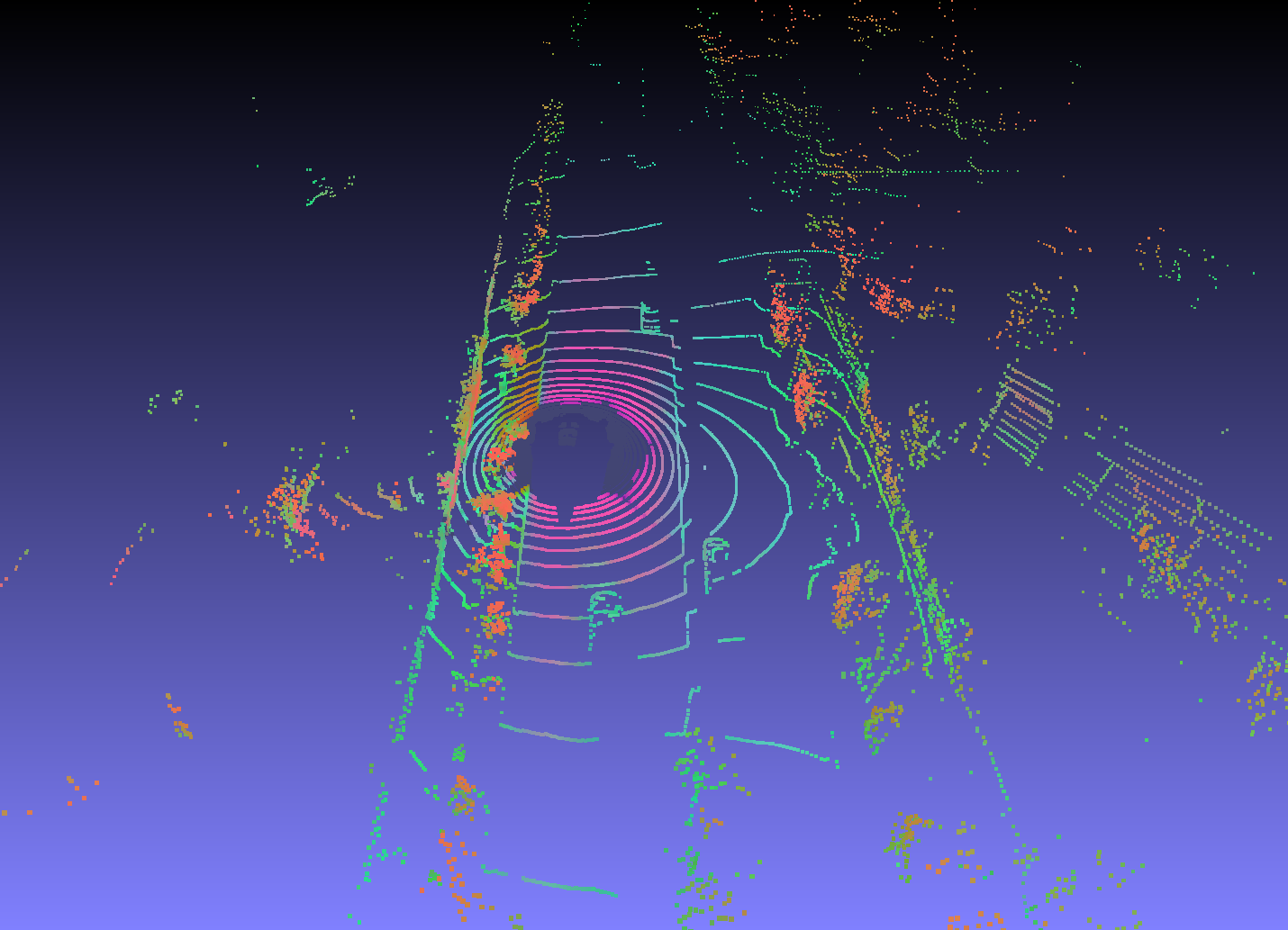} &
\includegraphics[width=0.31\linewidth]{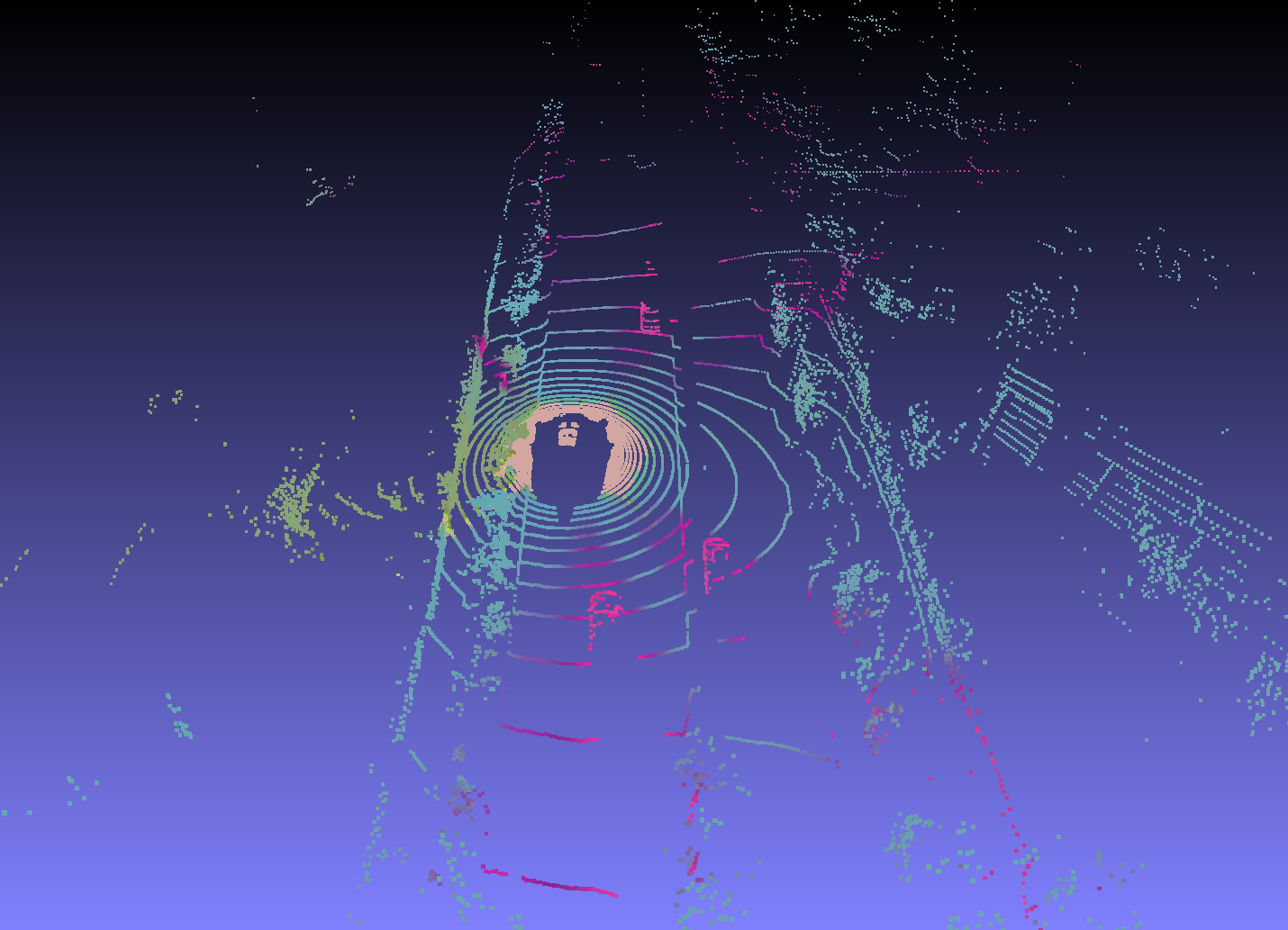} \\
\end{tabular}
\caption{Principal-component visualization of decoder level 2 U-Net
features on a representative nuScenes validation scene. Per-point feature
vectors are projected to three principal components and mapped to RGB.
The same scene under different task prompts produces visually distinct
feature structures, visualizing the cross-modal feature decomposition
quantified in Section~\ref{sec:cosine}. Additional scenes in
Appendix~\ref{app:qualitative_extended}.}
\label{fig:qualitative_pca}
\end{figure}

Figure~\ref{fig:qualitative_pca} visualizes U-Net decoder features at level 2, with per-point feature vectors reduced to their first three principal components and mapped to RGB. We select this level for its combination of low cross-modal cosine similarity and high linear-probe \gls{miou}.
\begin{figure}
\centering
\setlength{\tabcolsep}{2pt}
\renewcommand{\arraystretch}{0.6}
\begin{tabular}{@{}cc@{}}
\small Semantic Prompt & \small Instance Prompt \\
\includegraphics[width=0.45\linewidth]{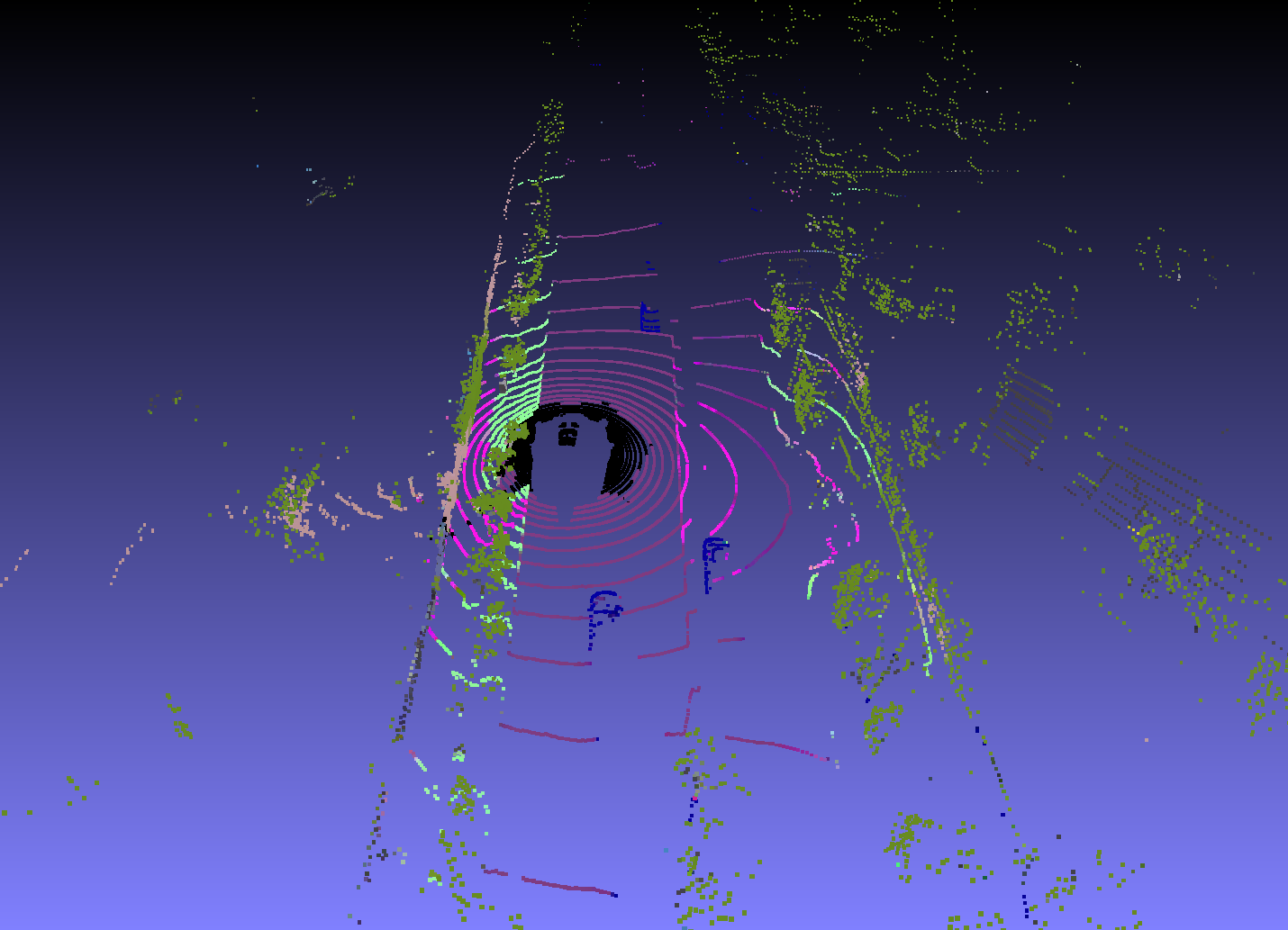} &
\includegraphics[width=0.45\linewidth]{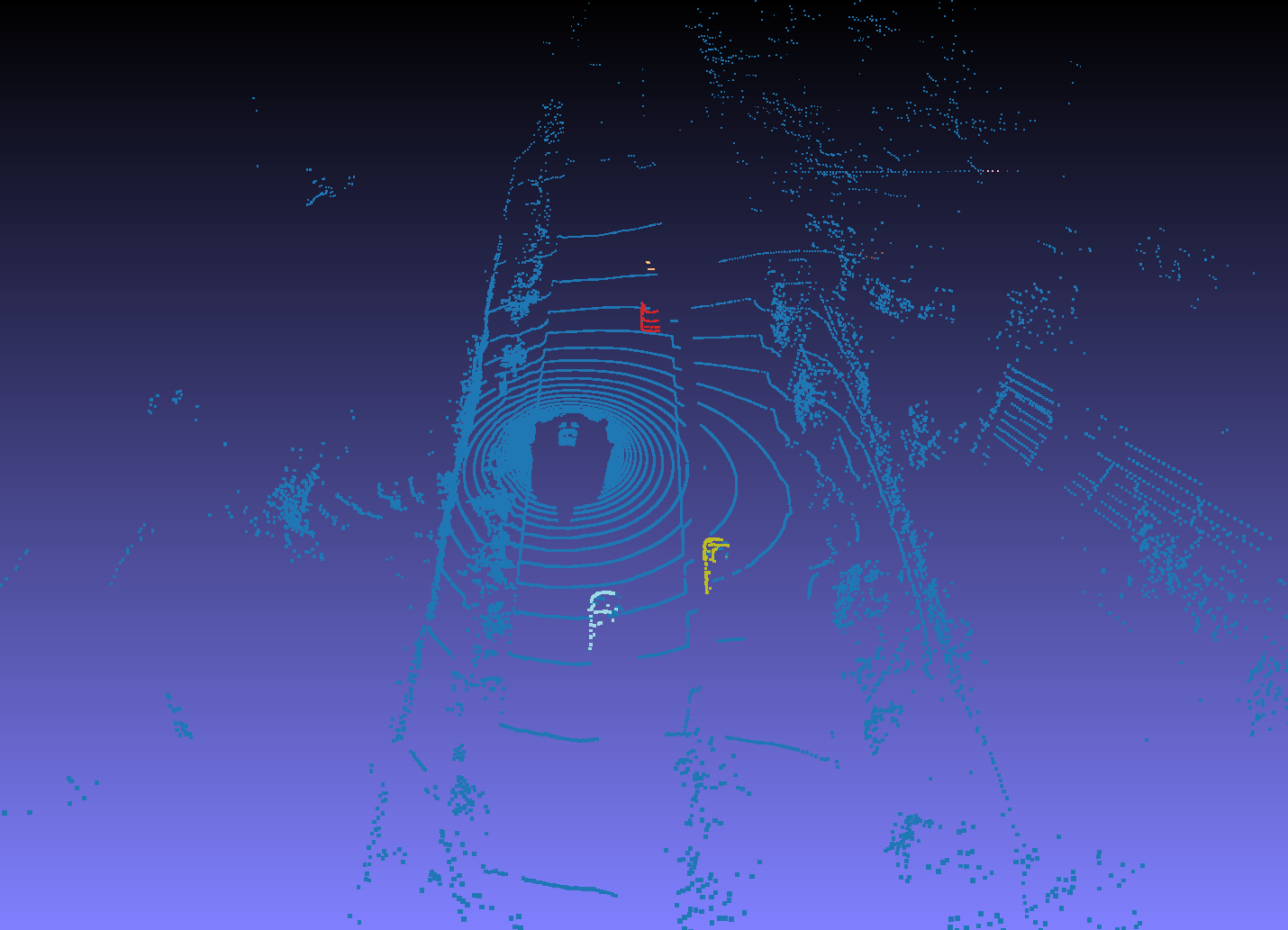}
\end{tabular}
\caption{Predicted outputs in point-cloud space on a representative nuScenes validation scene. Semantic outputs are color mapped, instance clusters are processed according to Section~\ref{sec:instance-seg}.}
\label{fig:qualitative_outputs}
\end{figure}
\section{Discussion and Limitations} \label{sec:discussion}
We have shown that a single LiDAR-conditioned diffusion model, trained entirely from 2D foundation-model pseudo-labels, learns features that encode non-trivial 3D structure and exhibit a layered cross-modal organization when probed in point-cloud space. The learned representation progresses from modality-specific shallow features, through a shared bottleneck, and back toward task-specialized decoder representations, suggesting that diffusion can act as a viable mechanism for transferring large-scale 2D priors into 3D. More broadly, this extends the diffusion-features-for-perception line of work~\cite{diff-features1, diff-features3, diff-features5} from single-modality 2D settings to a multi-modal, 3D-conditioned one.
Our analysis has several limitations. Experiments are conducted on a single dataset (nuScenes) and reported from a single training run, so cross-dataset generalization and sensitivity to random initialization remain untested. Quantitative semantic evaluation is constrained by the
label-space mismatch between the Cityscapes and nuScenes lidarseg taxonomies, while instance predictions are evaluated qualitatively since instance metrics depend strongly on a post-processing pipeline that is orthogonal to the representational analysis studied here. Output quality is not competitive with task-specialized 3D networks, as the objective of
this work is representation analysis rather than task-specific output
optimization. We therefore treat outputs primarily as probes of the learned features rather than as standalone benchmark targets.
Several directions follow naturally from this work. The same probing protocol could be applied to alternative LiDAR-conditioned diffusion architectures, 3D-aware transformers, or non-diffusion multi-task models, providing a broader picture of how structured cross-modal representations emerge. Downstream evaluation, for example, training lightweight task heads for 3D detection or panoptic segmentation on top of the lifted features, would test whether the recovered structure is operationally useful. Extending the analysis to additional modalities (optical flow, surface normals, motion forecasting) would test whether the layered organization is specific to the three modalities studied here or a general property of multi-modal diffusion training.
\begin{ack}
The research leading to these results is funded by the German Federal Ministry for Economic Affairs and Energy within the project “NXT GEN AI METHODS – Generative Methoden für Perzeption, Prädiktion und Planung" (grant no. 19A23014M).
\end{ack}

\newpage
\bibliographystyle{abbrvnat}
\small
\bibliography{references}

@INPROCEEDINGS{nuscenes,
  title={nuScenes: A multimodal dataset for autonomous driving},
  author={Holger Caesar and Varun Bankiti and Alex H. Lang and Sourabh Vora and 
          Venice Erin Liong and Qiang Xu and Anush Krishnan and Yu Pan and 
          Giancarlo Baldan and Oscar Beijbom}, 
  booktitle={CVPR},
  year=2020
}

@InProceedings{waymo, 
    author = {Sun, Pei and Kretzschmar, Henrik and Dotiwalla, Xerxes and Chouard, Aurelien and Patnaik, Vijaysai and Tsui, Paul and Guo, James and Zhou, Yin and Chai, Yuning and Caine, Benjamin and Vasudevan, Vijay and Han, Wei and Ngiam, Jiquan and Zhao, Hang and Timofeev, Aleksei and Ettinger, Scott and Krivokon, Maxim and Gao, Amy and Joshi, Aditya and Zhang, Yu and Shlens, Jonathon and Chen, Zhifeng and Anguelov, Dragomir}, 
    title = {Scalability in Perception for Autonomous Driving: Waymo Open Dataset},
    booktitle = {Proceedings of the IEEE/CVF Conference on Computer Vision and Pattern Recognition (CVPR)}, 
    month = {June}, 
    year = {2020} 
}

@article{kitti-raw,
  author = {Andreas Geiger and Philip Lenz and Christoph Stiller and Raquel Urtasun},
  title = {Vision meets Robotics: The KITTI Dataset},
  journal = {International Journal of Robotics Research (IJRR)},
  year = {2013}
}

@inproceedings{cityscapes,
  title={The cityscapes dataset for semantic urban scene understanding},
  author={Cordts, Marius and Omran, Mohamed and Ramos, Sebastian and Rehfeld, Timo and Enzweiler, Markus and Benenson, Rodrigo and Franke, Uwe and Roth, Stefan and Schiele, Bernt},
  booktitle={Proceedings of the IEEE conference on computer vision and pattern recognition},
  pages={3213--3223},
  year={2016}
}

@inproceedings{camera1-magicdrive,
 author = {Gao, Ruiyuan and Chen, Kai and Xie, Enze and HONG, Lanqing and Li, Zhenguo and Yeung, Dit-Yan and Xu, Qiang},
 booktitle = {International Conference on Learning Representations},
 editor = {B. Kim and Y. Yue and S. Chaudhuri and K. Fragkiadaki and M. Khan and Y. Sun},
 pages = {22841--22860},
 title = {MagicDrive: Street View Generation with Diverse 3D Geometry Control},
 url = {https://proceedings.iclr.cc/paper_files/paper/2024/file/635351dc3d18116d1a44ca31592e2f9a-Paper-Conference.pdf},
 volume = {2024},
 year = {2024}
}

@InProceedings{camera2-magicdrive2,
    author    = {Gao, Ruiyuan and Chen, Kai and Xiao, Bo and Hong, Lanqing and Li, Zhenguo and Xu, Qiang},
    title     = {MagicDrive-V2: High-Resolution Long Video Generation for Autonomous Driving with Adaptive Control},
    booktitle = {Proceedings of the IEEE/CVF International Conference on Computer Vision (ICCV)},
    month     = {October},
    year      = {2025},
    pages     = {28135-28144}
}

@article{camera3-bevcontrol,
  title={Bevcontrol: Accurately controlling street-view elements with multi-perspective consistency via bev sketch layout},
  author={Yang, Kairui and Ma, Enhui and Peng, Jibin and Guo, Qing and Lin, Di and Yu, Kaicheng},
  journal={arXiv preprint arXiv:2308.01661},
  year={2023}
}

@inproceedings{camera4-drivedreamer,
author = {Wang, Xiaofeng and Zhu, Zheng and Huang, Guan and Chen, Xinze and Zhu, Jiagang and Lu, Jiwen},
title = {DriveDreamer: Towards Real-World-Drive World Models for Autonomous Driving},
year = {2024},
isbn = {978-3-031-73194-5},
publisher = {Springer-Verlag},
address = {Berlin, Heidelberg},
url = {https://doi.org/10.1007/978-3-031-73195-2_4},
doi = {10.1007/978-3-031-73195-2_4},
booktitle = {Computer Vision – ECCV 2024: 18th European Conference, Milan, Italy, September 29–October 4, 2024, Proceedings, Part XLVIII},
pages = {55–72},
numpages = {18},
location = {Milan, Italy}
}

@article{camera5-vista,
  title={Vista: A generalizable driving world model with high fidelity and versatile controllability},
  author={Gao, Shenyuan and Yang, Jiazhi and Chen, Li and Chitta, Kashyap and Qiu, Yihang and Geiger, Andreas and Zhang, Jun and Li, Hongyang},
  journal={Advances in Neural Information Processing Systems},
  volume={37},
  pages={91560--91596},
  year={2024}
}

@inproceedings{camera6-panacea,
  title={Panacea: Panoramic and controllable video generation for autonomous driving},
  author={Wen, Yuqing and Zhao, Yucheng and Liu, Yingfei and Jia, Fan and Wang, Yanhui and Luo, Chong and Zhang, Chi and Wang, Tiancai and Sun, Xiaoyan and Zhang, Xiangyu},
  booktitle={Proceedings of the IEEE/CVF Conference on Computer Vision and Pattern Recognition},
  pages={6902--6912},
  year={2024}
}

@inproceedings{lc1-xdrive,
  title={X-Drive: Cross-modality Consistent Multi-Sensor Data Synthesis for Driving Scenarios},
  author={Xie, Yichen and Xu, Chenfeng and Peng, Chensheng and Zhao, Shuqi and Ho, Nhat and Pham, Alexander T and Ding, Mingyu and Tomizuka, Masayoshi and Zhan, Wei},
  booktitle={The Thirteenth International Conference on Learning Representations},
  year={2025}
}

@article{lc2-holodrive,
  title={Holodrive: Holistic 2d-3d multi-modal street scene generation for autonomous driving},
  author={Wu, Zehuan and Ni, Jingcheng and Wang, Xiaodong and Guo, Yuxin and Chen, Rui and Lu, Lewei and Dai, Jifeng and Xiong, Yuwen},
  journal={arXiv preprint arXiv:2412.01407},
  year={2024}
}

@inproceedings{lc3-uniscene,
  title={Uniscene: Unified occupancy-centric driving scene generation},
  author={Li, Bohan and Guo, Jiazhe and Liu, Hongsi and Zou, Yingshuang and Ding, Yikang and Chen, Xiwu and Zhu, Hu and Tan, Feiyang and Zhang, Chi and Wang, Tiancai and others},
  booktitle={Proceedings of the computer vision and pattern recognition conference},
  pages={11971--11981},
  year={2025}
}

@article{lc4-bevworld,
  title={BEVWorld: A multimodal world simulator for autonomous driving via scene-level BEV latents},
  author={Zhang, Yumeng and Gong, Shi and Xiong, Kaixin and Ye, Xiaoqing and Li, Xiaofan and Tan, Xiao and Wang, Fan and Huang, Jizhou and Wu, Hua and Wang, Haifeng},
  journal={arXiv preprint arXiv:2407.05679},
  year={2024}
}

@INPROCEEDINGS{lidar1-lidarDM,
  author={Zyrianov, Vlas and Che, Henry and Liu, Zhijian and Wang, Shenlong},
  booktitle={2025 IEEE International Conference on Robotics and Automation (ICRA)}, 
  title={LidarDM: Generative LiDAR Simulation in a Generated World}, 
  year={2025},
  volume={},
  number={},
  pages={6055-6062},
  doi={10.1109/ICRA55743.2025.11128001}}

@inproceedings{lidar2-r2dm,
  title={Lidar data synthesis with denoising diffusion probabilistic models},
  author={Nakashima, Kazuto and Kurazume, Ryo},
  booktitle={2024 IEEE International Conference on Robotics and Automation (ICRA)},
  pages={14724--14731},
  year={2024},
  organization={IEEE}
}

@inproceedings{lidar3-lidargen,
  title={Learning to generate realistic lidar point clouds},
  author={Zyrianov, Vlas and Zhu, Xiyue and Wang, Shenlong},
  booktitle={European Conference on Computer Vision},
  pages={17--35},
  year={2022},
  organization={Springer}
}

@inproceedings{lidar4-upsamling,
  title={Fast lidar upsampling using conditional diffusion models},
  author={Helgesen, Sander Elias Magnussen and Nakashima, Kazuto and T{\o}rresen, Jim and Kurazume, Ryo},
  booktitle={2024 33rd IEEE International Conference on Robot and Human Interactive Communication (ROMAN)},
  pages={272--277},
  year={2024},
  organization={IEEE}
}

@inproceedings{lidar5-LiDM,
  title={Towards realistic scene generation with lidar diffusion models},
  author={Ran, Haoxi and Guizilini, Vitor and Wang, Yue},
  booktitle={Proceedings of the IEEE/CVF Conference on Computer Vision and Pattern Recognition},
  pages={14738--14748},
  year={2024}
}

@inproceedings{lidar6-rangeldm,
  title={Rangeldm: Fast realistic lidar point cloud generation},
  author={Hu, Qianjiang and Zhang, Zhimin and Hu, Wei},
  booktitle={European Conference on Computer Vision},
  pages={115--135},
  year={2024},
  organization={Springer}
}

@inproceedings{lidar7-text2lidar,
  title={Text2lidar: Text-guided lidar point cloud generation via equirectangular transformer},
  author={Wu, Yang and Zhang, Kaihua and Qian, Jianjun and Xie, Jin and Yang, Jian},
  booktitle={European Conference on Computer Vision},
  pages={291--310},
  year={2024},
  organization={Springer}
}

@article{lidar8-lidarcrafter,
  title={LiDARCrafter: Dynamic 4D world modeling from LiDAR sequences},
  author={Liang, Ao and Liu, Youquan and Yang, Yu and Lu, Dongyue and Li, Linfeng and Kong, Lingdong and Zhao, Huaici and Ooi, Wei Tsang},
  journal={arXiv preprint arXiv:2508.03692},
  year={2025}
}

@article{diffusion1-ddpm,
  title={Denoising diffusion probabilistic models},
  author={Ho, Jonathan and Jain, Ajay and Abbeel, Pieter},
  journal={Advances in neural information processing systems},
  volume={33},
  pages={6840--6851},
  year={2020}
}

@inproceedings{diffusion2-ldm,
  title={High-resolution image synthesis with latent diffusion models},
  author={Rombach, Robin and Blattmann, Andreas and Lorenz, Dominik and Esser, Patrick and Ommer, Bj{\"o}rn},
  booktitle={Proceedings of the IEEE/CVF conference on computer vision and pattern recognition},
  pages={10684--10695},
  year={2022}
}

@inproceedings{diffusion3-earlywork,
  title={Deep unsupervised learning using nonequilibrium thermodynamics},
  author={Sohl-Dickstein, Jascha and Weiss, Eric and Maheswaranathan, Niru and Ganguli, Surya},
  booktitle={International conference on machine learning},
  pages={2256--2265},
  year={2015},
  organization={pmlr}
}

@inproceedings{diffusion5-controlnet,
  title={Adding conditional control to text-to-image diffusion models},
  author={Zhang, Lvmin and Rao, Anyi and Agrawala, Maneesh},
  booktitle={Proceedings of the IEEE/CVF international conference on computer vision},
  pages={3836--3847},
  year={2023}
}

@article{diffusion6-t2i, 
    title={T2I-Adapter: Learning Adapters to Dig Out More Controllable Ability for Text-to-Image Diffusion Models}, 
    volume={38}, 
    url={https://ojs.aaai.org/index.php/AAAI/article/view/28226}, 
    DOI={10.1609/aaai.v38i5.28226}, 
    number={5}, 
    journal={Proceedings of the AAAI Conference on Artificial Intelligence}, 
    author={Mou, Chong and Wang, Xintao and Xie, Liangbin and Wu, Yanze and Zhang, Jian and Qi, Zhongang and Shan, Ying}, 
    year={2024}, 
    month={Mar.},
    pages={4296-4304} }

@article{diffusion7-ipadapter,
  title={Ip-adapter: Text compatible image prompt adapter for text-to-image diffusion models},
  author={Ye, Hu and Zhang, Jun and Liu, Sibo and Han, Xiao and Yang, Wei},
  journal={arXiv preprint arXiv:2308.06721},
  year={2023}
}

@inproceedings{diffusion8-t2i-gligen,
  title={Gligen: Open-set grounded text-to-image generation},
  author={Li, Yuheng and Liu, Haotian and Wu, Qingyang and Mu, Fangzhou and Yang, Jianwei and Gao, Jianfeng and Li, Chunyuan and Lee, Yong Jae},
  booktitle={Proceedings of the IEEE/CVF conference on computer vision and pattern recognition},
  pages={22511--22521},
  year={2023}
}

@InProceedings{diffusion9-t2i-glide,
  title = 	 {{GLIDE}: Towards Photorealistic Image Generation and Editing with Text-Guided Diffusion Models},
  author =       {Nichol, Alexander Quinn and Dhariwal, Prafulla and Ramesh, Aditya and Shyam, Pranav and Mishkin, Pamela and Mcgrew, Bob and Sutskever, Ilya and Chen, Mark},
  booktitle = 	 {Proceedings of the 39th International Conference on Machine Learning},
  pages = 	 {16784--16804},
  year = 	 {2022},
  editor = 	 {Chaudhuri, Kamalika and Jegelka, Stefanie and Song, Le and Szepesvari, Csaba and Niu, Gang and Sabato, Sivan},
  volume = 	 {162},
  series = 	 {Proceedings of Machine Learning Research},
  month = 	 {17--23 Jul},
  publisher =    {PMLR},
  url = 	 {https://proceedings.mlr.press/v162/nichol22a.html}
}

@article{diffusion10-t2i-imagen,
  title={Photorealistic text-to-image diffusion models with deep language understanding},
  author={Saharia, Chitwan and Chan, William and Saxena, Saurabh and Li, Lala and Whang, Jay and Denton, Emily L and Ghasemipour, Kamyar and Gontijo Lopes, Raphael and Karagol Ayan, Burcu and Salimans, Tim and others},
  journal={Advances in neural information processing systems},
  volume={35},
  pages={36479--36494},
  year={2022}
}

@article{diffusion11-dalle,
  title={Hierarchical Text-Conditional Image Generation with CLIP Latents},
  author={Ramesh, Aditya and Dhariwal, Prafulla and Nichol, Alex and Chu, Casey and Chen, Mark},
  journal={arXiv e-prints},
  pages={arXiv--2204},
  year={2022}
}

@inproceedings{diffusion12-cfg,
title={Classifier-Free Diffusion Guidance},
author={Jonathan Ho and Tim Salimans},
booktitle={NeurIPS 2021 Workshop on Deep Generative Models and Downstream Applications},
year={2021},
url={https://openreview.net/forum?id=qw8AKxfYbI}
}

@inproceedings{diffusion13-depth,
  title={Repurposing diffusion-based image generators for monocular depth estimation},
  author={Ke, Bingxin and Obukhov, Anton and Huang, Shengyu and Metzger, Nando and Daudt, Rodrigo Caye and Schindler, Konrad},
  booktitle={Proceedings of the IEEE/CVF conference on computer vision and pattern recognition},
  pages={9492--9502},
  year={2024}
}

@inproceedings{diffusion14-segmentation,
  title={Open-vocabulary panoptic segmentation with text-to-image diffusion models},
  author={Xu, Jiarui and Liu, Sifei and Vahdat, Arash and Byeon, Wonmin and Wang, Xiaolong and De Mello, Shalini},
  booktitle={Proceedings of the IEEE/CVF conference on computer vision and pattern recognition},
  pages={2955--2966},
  year={2023}
}

@article{diffusion15-flow,
  title={The surprising effectiveness of diffusion models for optical flow and monocular depth estimation},
  author={Saxena, Saurabh and Herrmann, Charles and Hur, Junhwa and Kar, Abhishek and Norouzi, Mohammad and Sun, Deqing and Fleet, David J},
  journal={Advances in Neural Information Processing Systems},
  volume={36},
  pages={39443--39469},
  year={2023}
}

@inproceedings{diffusion16-light,
author = {Magar, Nadav and Hertz, Amir and Tabellion, Eric and Pritch, Yael and Rav-Acha, Alex and Shamir, Ariel and Hoshen, Yedid},
title = {LightLab: Controlling Light Sources in Images with Diffusion Models},
year = {2025},
isbn = {9798400715402},
publisher = {Association for Computing Machinery},
address = {New York, NY, USA},
url = {https://doi.org/10.1145/3721238.3730696},
doi = {10.1145/3721238.3730696},
booktitle = {Proceedings of the Special Interest Group on Computer Graphics and Interactive Techniques Conference Conference Papers},
articleno = {106},
numpages = {11},
location = {
},
series = {SIGGRAPH Conference Papers '25}
}

@article{diffusion17-spatial1,
  title={Denoising diffusion restoration models},
  author={Kawar, Bahjat and Elad, Michael and Ermon, Stefano and Song, Jiaming},
  journal={Advances in neural information processing systems},
  volume={35},
  pages={23593--23606},
  year={2022}
}

@article{diffusion18-spatial2,
  title={Exploiting diffusion prior for real-world image super-resolution},
  author={Wang, Jianyi and Yue, Zongsheng and Zhou, Shangchen and Chan, Kelvin CK and Loy, Chen Change},
  journal={International Journal of Computer Vision},
  volume={132},
  number={12},
  pages={5929--5949},
  year={2024},
  publisher={Springer}
}

@inproceedings{diffusion19-spatial3,
  title={Palette: Image-to-image diffusion models},
  author={Saharia, Chitwan and Chan, William and Chang, Huiwen and Lee, Chris and Ho, Jonathan and Salimans, Tim and Fleet, David and Norouzi, Mohammad},
  booktitle={ACM SIGGRAPH 2022 conference proceedings},
  pages={1--10},
  year={2022}
}

@article{foundation1-dinov2,
  title={DINOv2: Learning Robust Visual Features without Supervision},
  author={Oquab, Maxime and Darcet, Timoth{\'e}e and Moutakanni, Th{\'e}o and Vo, Huy and Szafraniec, Marc and Khalidov, Vasil and Fernandez, Pierre and Haziza, Daniel and Massa, Francisco and El-Nouby, Alaaeldin and others},
  journal={Transactions on Machine Learning Research Journal},
  pages={1--31},
  year={2024}
}

@article{foundation2-dinov3,
  title={Dinov3},
  author={Sim{\'e}oni, Oriane and Vo, Huy V and Seitzer, Maximilian and Baldassarre, Federico and Oquab, Maxime and Jose, Cijo and Khalidov, Vasil and Szafraniec, Marc and Yi, Seungeun and Ramamonjisoa, Micha{\"e}l and others},
  journal={arXiv preprint arXiv:2508.10104},
  year={2025}
}

@article{foundation3-depthanything3,
  title={Depth Anything 3: Recovering the visual space from any views},
  author={Haotong Lin and Sili Chen and Jun Hao Liew and Donny Y. Chen and Zhenyu Li and Guang Shi and Jiashi Feng and Bingyi Kang},
  journal={arXiv preprint arXiv:2511.10647},
  year={2025}
}

@article{foundation4-depthanything2,
  title={Depth anything v2},
  author={Yang, Lihe and Kang, Bingyi and Huang, Zilong and Zhao, Zhen and Xu, Xiaogang and Feng, Jiashi and Zhao, Hengshuang},
  journal={Advances in Neural Information Processing Systems},
  volume={37},
  pages={21875--21911},
  year={2024}
}

@inproceedings{foundation5-sam2,
title={{SAM} 2: Segment Anything in Images and Videos},
author={Nikhila Ravi and Valentin Gabeur and Yuan-Ting Hu and Ronghang Hu and Chaitanya Ryali and Tengyu Ma and Haitham Khedr and Roman R{\"a}dle and Chloe Rolland and Laura Gustafson and Eric Mintun and Junting Pan and Kalyan Vasudev Alwala and Nicolas Carion and Chao-Yuan Wu and Ross Girshick and Piotr Dollar and Christoph Feichtenhofer},
booktitle={The Thirteenth International Conference on Learning Representations},
year={2025},
}

@inproceedings{foundation6-sam3,
title={{SAM} 3: Segment Anything with Concepts},
author={Nicolas Carion and Laura Gustafson and Yuan-Ting Hu and Shoubhik Debnath and Ronghang Hu and Didac Suris Coll-Vinent and Chaitanya Ryali and Kalyan Vasudev Alwala and Haitham Khedr and Andrew Huang and Jie Lei and Tengyu Ma and Baishan Guo and Arpit Kalla and Markus Marks and Joseph Greer and Meng Wang and Peize Sun and Roman R{\"a}dle and Triantafyllos Afouras and Effrosyni Mavroudi and Katherine Xu and Tsung-Han Wu and Yu Zhou and Liliane Momeni and RISHI HAZRA and Shuangrui Ding and Sagar Vaze and Francois Porcher and Feng Li and Siyuan Li and Aishwarya Kamath and Ho Kei Cheng and Piotr Dollar and Nikhila Ravi and Kate Saenko and Pengchuan Zhang and Christoph Feichtenhofer},
booktitle={The Fourteenth International Conference on Learning Representations},
year={2026},
url={https://openreview.net/forum?id=r35clVtGzw}
}

@inproceedings{foundation11-dino,
  title={Emerging properties in self-supervised vision transformers},
  author={Caron, Mathilde and Touvron, Hugo and Misra, Ishan and J{\'e}gou, Herv{\'e} and Mairal, Julien and Bojanowski, Piotr and Joulin, Armand},
  booktitle={Proceedings of the IEEE/CVF international conference on computer vision},
  pages={9650--9660},
  year={2021}
}

@inproceedings{panoptic-deeplab,
  title={Panoptic-deeplab: A simple, strong, and fast baseline for bottom-up panoptic segmentation},
  author={Cheng, Bowen and Collins, Maxwell D and Zhu, Yukun and Liu, Ting and Huang, Thomas S and Adam, Hartwig and Chen, Liang-Chieh},
  booktitle={Proceedings of the IEEE/CVF conference on computer vision and pattern recognition},
  pages={12475--12485},
  year={2020}
}

@article{depth-least-squares,
  title={Towards robust monocular depth estimation: Mixing datasets for zero-shot cross-dataset transfer},
  author={Ranftl, Ren{\'e} and Lasinger, Katrin and Hafner, David and Schindler, Konrad and Koltun, Vladlen},
  journal={IEEE transactions on pattern analysis and machine intelligence},
  volume={44},
  number={3},
  pages={1623--1637},
  year={2020},
  publisher={IEEE}
}

@inproceedings{depth1-eval,
  title={Vision transformers for dense prediction},
  author={Ranftl, Ren{\'e} and Bochkovskiy, Alexey and Koltun, Vladlen},
  booktitle={Proceedings of the IEEE/CVF international conference on computer vision},
  pages={12179--12188},
  year={2021}
}

@inproceedings{depth2-eval,
  title={Metric3d: Towards zero-shot metric 3d prediction from a single image},
  author={Yin, Wei and Zhang, Chi and Chen, Hao and Cai, Zhipeng and Yu, Gang and Wang, Kaixuan and Chen, Xiaozhi and Shen, Chunhua},
  booktitle={Proceedings of the IEEE/CVF international conference on computer vision},
  pages={9043--9053},
  year={2023}
}

@inproceedings{depth3-eval,
  title={Geowizard: Unleashing the diffusion priors for 3d geometry estimation from a single image},
  author={Fu, Xiao and Yin, Wei and Hu, Mu and Wang, Kaixuan and Ma, Yuexin and Tan, Ping and Shen, Shaojie and Lin, Dahua and Long, Xiaoxiao},
  booktitle={European Conference on Computer Vision},
  pages={241--258},
  year={2024},
  organization={Springer}
}

@inproceedings{depth4-eval,
  title={Depthfm: Fast generative monocular depth estimation with flow matching},
  author={Gui, Ming and Schusterbauer, Johannes and Prestel, Ulrich and Ma, Pingchuan and Kotovenko, Dmytro and Grebenkova, Olga and Baumann, Stefan Andreas and Hu, Vincent Tao and Ommer, Bj{\"o}rn},
  booktitle={Proceedings of the AAAI Conference on Artificial Intelligence},
  volume={39},
  number={3},
  pages={3203--3211},
  year={2025}
}

@inproceedings{wu2025sonata,
  title={Sonata: Self-supervised learning of reliable point representations},
  author={Wu, Xiaoyang and DeTone, Daniel and Frost, Duncan and Shen, Tianwei and Xie, Chris and Yang, Nan and Engel, Jakob and Newcombe, Richard and Zhao, Hengshuang and Straub, Julian},
  booktitle={Proceedings of the Computer Vision and Pattern Recognition Conference},
  pages={22193--22204},
  year={2025}
}

@article{zhang2026utonia,
  title={Utonia: Toward One Encoder for All Point Clouds},
  author={Zhang, Yujia and Wu, Xiaoyang and Yang, Yunhan and Fan, Xianzhe and Li, Han and Zhang, Yuechen and Huang, Zehao and Wang, Naiyan and Zhao, Hengshuang},
  journal={arXiv preprint arXiv:2603.03283},
  year={2026}
}

@inproceedings{zhou2023uni3d,
  title={Uni3d: Exploring unified 3d representation at scale},
  author={Zhou, Junsheng and Wang, Jinsheng and Ma, Baorui and Liu, Yu-Shen and Huang, Tiejun and Wang, Xinlong},
  booktitle={International Conference on Learning Representations (ICLR)},
  year={2024}
}

@article{liu2023openshape,
  title={Openshape: Scaling up 3d shape representation towards open-world understanding},
  author={Liu, Minghua and Shi, Ruoxi and Kuang, Kaiming and Zhu, Yinhao and Li, Xuanlin and Han, Shizhong and Cai, Hong and Porikli, Fatih and Su, Hao},
  journal={Advances in neural information processing systems},
  volume={36},
  pages={44860--44879},
  year={2023}
}

@inproceedings{diff-features1,
  title={Unleashing text-to-image diffusion models for visual perception},
  author={Zhao, Wenliang and Rao, Yongming and Liu, Zuyan and Liu, Benlin and Zhou, Jie and Lu, Jiwen},
  booktitle={Proceedings of the IEEE/CVF international conference on computer vision},
  pages={5729--5739},
  year={2023}
}

@InProceedings{diff-features2,
    author    = {Xu, Jiarui and Liu, Sifei and Vahdat, Arash and Byeon, Wonmin and Wang, Xiaolong and De Mello, Shalini},
    title     = {Open-Vocabulary Panoptic Segmentation With Text-to-Image Diffusion Models},
    booktitle = {Proceedings of the IEEE/CVF Conference on Computer Vision and Pattern Recognition (CVPR)},
    month     = {June},
    year      = {2023},
    pages     = {2955-2966}
}

@article{diff-features3,
  title={Emergent correspondence from image diffusion},
  author={Tang, Luming and Jia, Menglin and Wang, Qianqian and Phoo, Cheng Perng and Hariharan, Bharath},
  journal={Advances in neural information processing systems},
  volume={36},
  pages={1363--1389},
  year={2023}
}

@inproceedings{diff-features4,
  title={Cleandift: Diffusion features without noise},
  author={Stracke, Nick and Baumann, Stefan Andreas and Bauer, Kolja and Fundel, Frank and Ommer, Bj{\"o}rn},
  booktitle={Proceedings of the IEEE/CVF Conference on Computer Vision and Pattern Recognition},
  pages={117--127},
  year={2025}
}

@article{diff-features5,
  title={Diffusion hyperfeatures: Searching through time and space for semantic correspondence},
  author={Luo, Grace and Dunlap, Lisa and Park, Dong Huk and Holynski, Aleksander and Darrell, Trevor},
  journal={Advances in Neural Information Processing Systems},
  volume={36},
  pages={47500--47510},
  year={2023}
}

@article{diff-features6,
  title={Maskdiffusion: Exploiting pre-trained diffusion models for semantic segmentation},
  author={Kawano, Yasufumi and Aoki, Yoshimitsu},
  journal={IEEE Access},
  volume={12},
  pages={127283--127293},
  year={2024},
  publisher={IEEE}
}

@article{xie2021segformer,
  title={SegFormer: Simple and efficient design for semantic segmentation with transformers},
  author={Xie, Enze and Wang, Wenhai and Yu, Zhiding and Anandkumar, Anima and Alvarez, Jose M and Luo, Ping},
  journal={Advances in neural information processing systems},
  volume={34},
  pages={12077--12090},
  year={2021}
}

@inproceedings{dbscan,
author = {Ester, Martin and Kriegel, Hans-Peter and Sander, J\"{o}rg and Xu, Xiaowei},
title = {A density-based algorithm for discovering clusters in large spatial databases with noise},
year = {1996},
publisher = {AAAI Press},
booktitle = {Proceedings of the Second International Conference on Knowledge Discovery and Data Mining},
pages = {226–231},
numpages = {6},
location = {Portland, Oregon},
series = {KDD'96}
}

@article{adamw,
  title={Decoupled weight decay regularization},
  author={Loshchilov, Ilya and Hutter, Frank},
  journal={arXiv preprint arXiv:1711.05101},
  year={2017}
}

@inproceedings{instructpix2pix,
  title={Instructpix2pix: Learning to follow image editing instructions},
  author={Brooks, Tim and Holynski, Aleksander and Efros, Alexei A},
  booktitle={Proceedings of the IEEE/CVF conference on computer vision and pattern recognition},
  pages={18392--18402},
  year={2023}
}

\newpage
\appendix
\section{Preliminaries}
\subsection{Latent Diffusion Models}
\label{app:diffusion}
Diffusion probabilistic models define a forward noising process that
gradually perturbs data samples into Gaussian noise, and a learned
reverse process that reconstructs data from noise. Let
$x_0 \sim p_{\text{data}}(\mathbf{x})$ denote a data sample. The
forward diffusion process is defined as a fixed Markov chain:
\begin{equation}
    q(x_t|x_{t-1}) = \mathcal{N}(x_t; \sqrt{1-\beta_t}\,x_{t-1}, \beta_t I),
\end{equation}
where $\{\beta_t\}_{t=1}^T$ is a predefined variance schedule. With
$\alpha_t = 1 - \beta_t$ and $\bar\alpha_t = \prod_{s=1}^t \alpha_s$,
this yields the closed-form marginal:
\begin{equation}
    q(x_t|x_0) = \mathcal{N}(x_t; \sqrt{\bar\alpha_t}\,x_0, (1-\bar\alpha_t) I).
\end{equation}
A neural network $\epsilon_\theta(x_t, t, c)$ is trained to predict
the added noise under conditioning $c$. Given
$\epsilon \sim \mathcal{N}(0, I)$ and
$x_t = \sqrt{\bar\alpha_t}\,x_0 + \sqrt{1-\bar\alpha_t}\,\epsilon$,
the standard objective is
\begin{equation}
    \mathcal{L} = \mathbb{E}_{x_0, \epsilon, t}
    \left[\|\epsilon - \epsilon_\theta(x_t, t, c)\|_2^2\right].
\end{equation}
Latent diffusion models~\cite{diffusion2-ldm} reduce computational cost
by performing diffusion in a compressed latent space. An encoder
$\mathcal{E}$ maps the input to $z_0 = \mathcal{E}(x_0)$, where
diffusion is applied; a decoder $D$ reconstructs the output
$\hat{x}_0 = D(z_0)$. The training objective becomes
\begin{equation}
    \mathcal{L}_{\text{LDM}} = \mathbb{E}_{z_0, \epsilon, t}
    \left[\|\epsilon - \epsilon_\theta(z_t, t, c)\|_2^2\right].
\end{equation}
\subsection{LiDAR-Camera Projection}
\label{app:projection}
Let $\mathbf{p}_L = (x, y, z, 1)^\top$ be a LiDAR point in homogeneous
coordinates and $T_{CL} \in SE(3)$ the rigid transformation from LiDAR
to camera frame. The point in camera coordinates is
\begin{equation}
    \tilde{\mathbf{p}}_C = T_{CL}\,\mathbf{p}_L,
\end{equation}
with Euclidean form $\mathbf{p}_C = (X, Y, Z)^\top$. The projection
$\pi: \mathbb{R}^3 \to \mathbb{R}^2$ onto the image plane uses the
camera intrinsic $K \in \mathbb{R}^{3 \times 3}$:
\begin{equation}
    \lambda \begin{pmatrix} u \\ v \\ 1 \end{pmatrix} = K\,\mathbf{p}_C,
    \quad
    u = \frac{f_x X}{Z} + c_x,
    \quad
    v = \frac{f_y Y}{Z} + c_y,
\end{equation}
with $\lambda = Z$.
Given a 2D pixel $\mathbf{u} = (u, v)^\top$ at depth
$\lambda = D(u, v)$, the backprojection
$\pi^{-1}: \mathbb{R}^2 \times \mathbb{R}^1 \to \mathbb{R}^3$ is
\begin{equation}
    \mathbf{p}_C = \pi^{-1}(\mathbf{u}, \lambda)
    = \lambda \cdot K^{-1}\,\tilde{\mathbf{u}},
\end{equation}
with $\tilde{\mathbf{u}} = (u, v, 1)^\top$. The point in LiDAR frame
is recovered by
\begin{equation}
    \mathbf{p}_L = T_{CL}^{-1}\,\tilde{\mathbf{p}}_C.
\end{equation}
We denote the composite projection from a 3D LiDAR point to 2D pixel
coordinates as
$\Pi: SE(3) \times \mathbb{R}^4 \to \mathbb{R}^2$:
\begin{equation}
    \mathbf{u}_i = \Pi(T_{CL}, \mathbf{p}_{L,i})
    = \pi(T_{CL}\,\mathbf{p}_{L,i}),
\end{equation}
and the full backprojection from a pixel and depth as
$\Pi^{-1}: SE(3) \times \mathbb{R}^2 \times \mathbb{R}^1 \to \mathbb{R}^4$:
\begin{equation}
    \mathbf{p}_{L,i} = \Pi^{-1}(T_{CL}, \mathbf{u}_i, \lambda_i)
    = T_{CL}^{-1}
    \begin{pmatrix} \pi^{-1}(\mathbf{u}_i, \lambda_i) \\ 1 \end{pmatrix}.
\end{equation}
\section{Short-Probe Convergence}
\label{app:short_probes}
\begin{figure}
\centering
\includegraphics[width=\linewidth]{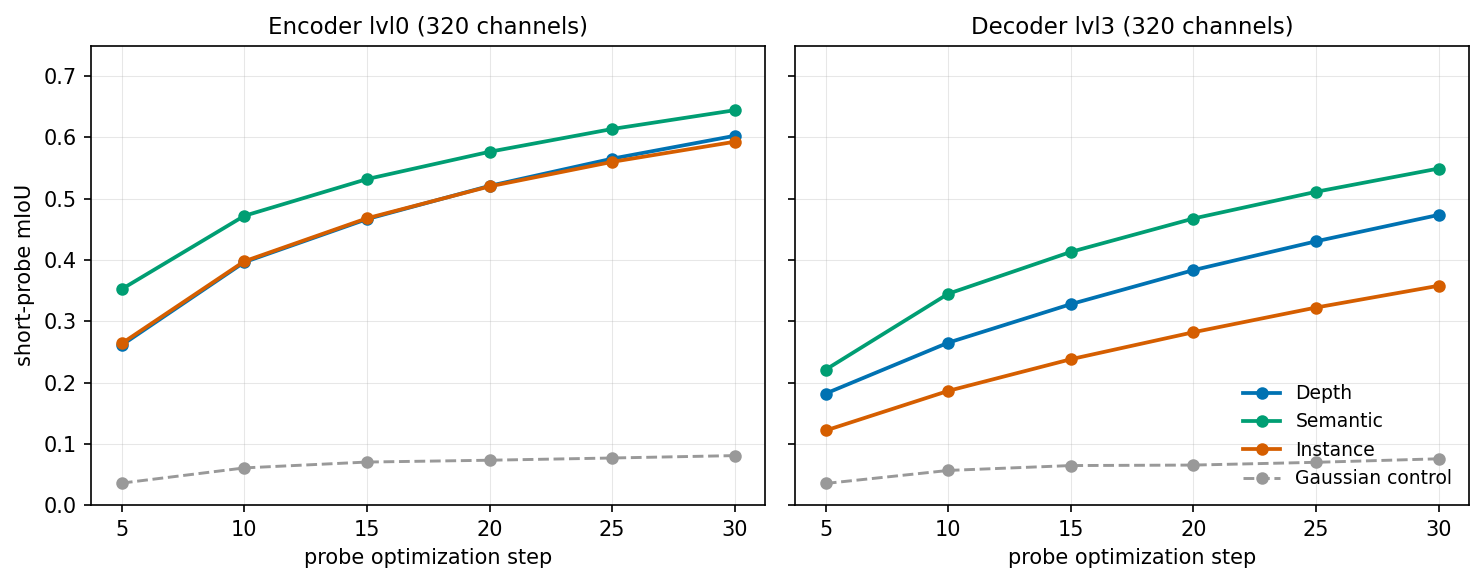}
\caption{Short-probe convergence at the two narrowest U-Net levels of Stable
Diffusion 1.5 (encoder level 0 and decoder level 3, both 320 channels), for
the three task-conditioned feature streams and the Gaussian-noise
control of matched dimensionality. At these levels the probe has too
few channels to fit noise: all three features reach $\sim$0.36--0.65
mIoU within 30 steps, while the control flattens near $\sim$0.08. The
gap reflects intrinsic separability rather than probe capacity.}
\label{fig:short_lowchannel}
\end{figure}

In addition to the long-range linear probes used for the main analysis
(Section~\ref{sec:probing}), we also compute a short-range probe
during training as a per-step diagnostic. The short probe is trained
for 30 optimization steps on a single sample (one batch of 6 views,
all three modalities), with intermediate scores logged every 5 steps.
Unlike the long probe, the short probe does not isolate generalization:
it is trained and evaluated on the same sample and primarily reflects
the linear separability of the features under a fixed optimization
budget. We use it as a training-monitoring signal, not as a measure of
representational quality.
Figure~\ref{fig:short_lowchannel} shows short-probe convergence at the
two narrowest U-Net levels in SD~1.5: encoder level 0 and decoder level 3,
both with 320 channels. At these low-dimensional levels, the linear probe
has limited freedom to overfit, so the comparison between real features
and the Gaussian-noise control is informative: a meaningful gap
indicates that the features themselves carry separable structure beyond
what the probe could synthesize from random vectors of matched
dimensionality. Across both panels, all three task-conditioned features
reach $\sim$0.36--0.65 \gls{miou} within 30 probe steps, while the
control flattens at $\sim$0.08. This $\sim 5\times$ gap, in a regime
where the probe has $4\times$ fewer channels to fit through than at the
1280-channel decoder level 0, is consistent with the long-probe finding
that diffusion features carry linearly decodable 3D structure.
\section{Implementation Details}
\label{app:implementation}
\subsection{Text Prompt Format}
\label{app:prompts}
We use a fixed two-field textual prompt of the form ``\texttt{Task: <modality>. Scene: <description>.}'', where \texttt{<modality>} is one of \{\texttt{depth},
\texttt{semantic\_segmentation}, \texttt{instance}\} and \texttt{<description>} is a short scene caption derived from the nuScenes scene metadata. The same scene description is used across
modalities for a given sample; only the \texttt{Task} field varies between the three modality-conditioned forward passes used for cross-modal cosine analysis (Section~\ref{sec:cosine}).
\subsection{Training Configuration}
\label{app:training}
We initialize the U-Net from Stable Diffusion 1.5~\cite{diffusion2-ldm}
and expand its input convolution to accept 6 channels: the original 4
VAE-latent channels concatenated with 2 conditioning channels (LiDAR
depth and intensity). The added input weights are zero-initialized
following~\cite{instructpix2pix}. All other weights inherit the
pretrained SD 1.5 values. Input images are processed at $224 \times 400$
resolution, yielding latents of shape $28 \times 50$. The sparse
LiDAR projection is max-pooled to the latent resolution.

We train with AdamW~\cite{adamw} ($\text{lr} = 4 \times 10^{-4}$, weight decay $0.01$) using a cosine schedule with $1000$ warmup steps and a minimum learning rate of
$10^{-5}$. Classifier-free guidance dropout and conditioning-projection dropout are both set to $20\%$.

All three task modalities are trained jointly: each batch is expanded along the modality axis so a single optimizer step sees all three prompts. With per-GPU batch size $4$, $6$ camera views per sample, $3$ modalities, and $8$ NVIDIA A40 GPUs, the effective batch size is
$576$. We train for $50$ epochs on nuScenes~\cite{nuscenes}.
\subsection{Probing Configuration}
\label{app:probing}
Both probes use a single linear layer mapping per-point feature vectors
to per-class logits, with no pooling and no nonlinearities. The probe
sees only the backprojected feature vector at each LiDAR point;
spatial coordinates $(x, y, z)$ are not provided as input. Probes are
trained with the cross-entropy loss against the pseudo-label class
assignment at each point.

The short-range probe is trained for $30$ optimization steps on a
single sample (one batch of $6$ views, all three modalities), with
intermediate scores logged every $5$ steps. The long-range probe is
trained for $60$ optimization steps. Each step accumulates features
and labels from a fixed validation chunk of $10$ samples and computes
the gradient over the concatenated set. We use full-batch updates
rather than stochastic minibatches. Evaluation is performed on a
disjoint $10$-sample bucket sampled from the unobserved validation
set, using mean intersection-over-union over the evaluable class set.

Both probes use AdamW with learning rate
$10^{-2}$ and weight decay $0$. Spatial coordinates are excluded by
construction: the input feature is the per-point backprojected U-Net
activation, with no positional encoding or coordinate concatenation.
The Gaussian-noise control feature used as a baseline is constructed
by computing the per-channel mean and standard deviation of the U-Net
features for a given level and sampling a tensor of matched shape from
$\mathcal{N}(\mu_c, \sigma_c)$.
\section{Additional Qualitative Results}
\label{app:qualitative_extended}
\begin{figure}
\centering
\setlength{\tabcolsep}{1pt}
\renewcommand{\arraystretch}{0.4}
\begin{tabular}{@{}cccc@{}}
& \small Front-Left & \small Front & \small Front-Right \\
\rotatebox{90}{\hspace{0.6em}\small Depth} &
\includegraphics[width=0.27\linewidth]{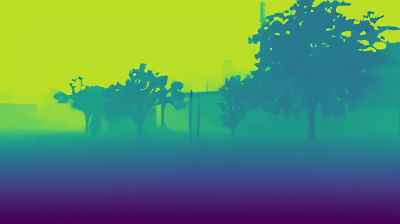} &
\includegraphics[width=0.27\linewidth]{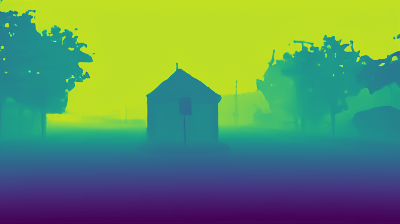} &
\includegraphics[width=0.27\linewidth]{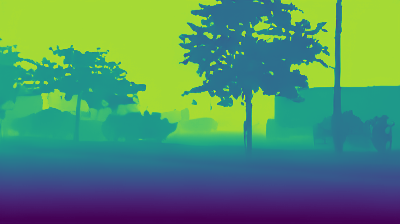} \\
\rotatebox{90}{\hspace{0.6em}\small Semantic} &
\includegraphics[width=0.27\linewidth]{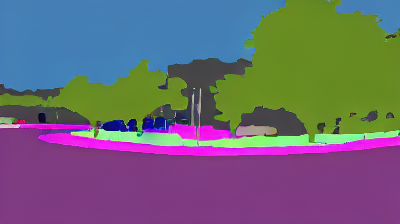} &
\includegraphics[width=0.27\linewidth]{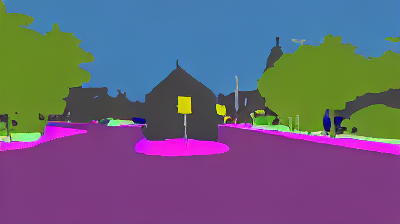} &
\includegraphics[width=0.27\linewidth]{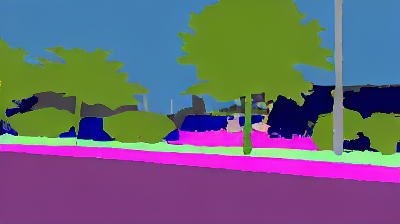} \\
\rotatebox{90}{\hspace{0.6em}\small Instance} &
\includegraphics[width=0.27\linewidth]{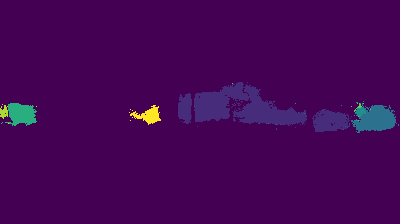} &
\includegraphics[width=0.27\linewidth]{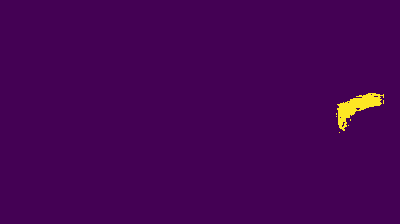} &
\includegraphics[width=0.27\linewidth]{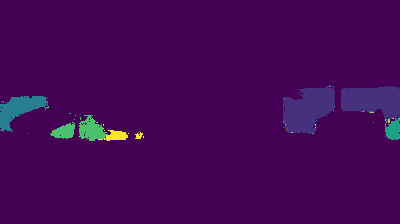} \\[6pt]
& \small Back-Left & \small Back & \small Back-Right \\
\rotatebox{90}{\hspace{0.6em}\small Depth} &
\includegraphics[width=0.27\linewidth]{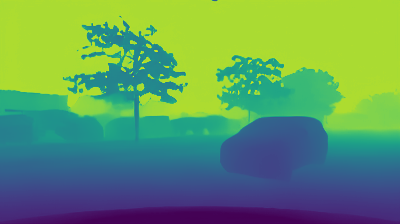} &
\includegraphics[width=0.27\linewidth]{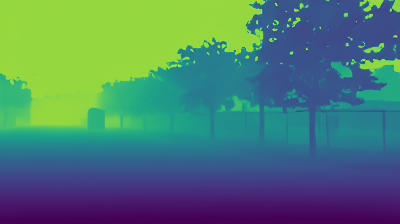} &
\includegraphics[width=0.27\linewidth]{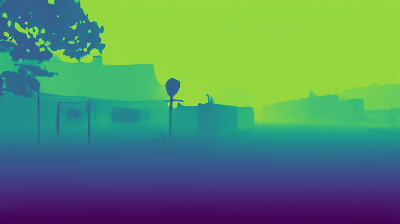} \\
\rotatebox{90}{\hspace{0.6em}\small Semantic} &
\includegraphics[width=0.27\linewidth]{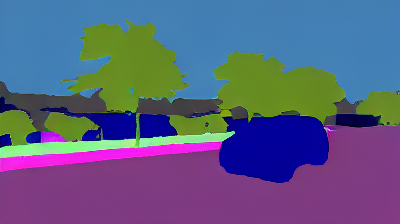} &
\includegraphics[width=0.27\linewidth]{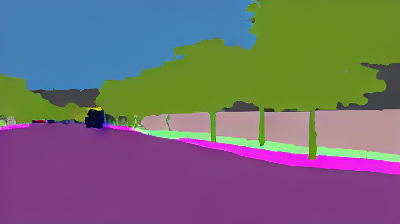} &
\includegraphics[width=0.27\linewidth]{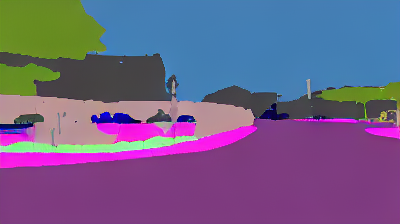} \\
\rotatebox{90}{\hspace{0.6em}\small Instance} &
\includegraphics[width=0.27\linewidth]{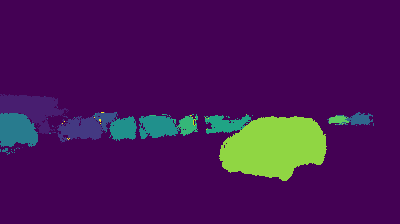} &
\includegraphics[width=0.27\linewidth]{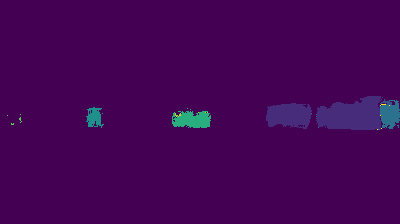} &
\includegraphics[width=0.27\linewidth]{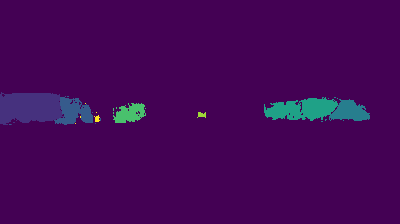} \\
\end{tabular}
\caption{Direct 2D outputs of the model across all six nuScenes camera
views. Top: front three views (FL/F/FR). Bottom: rear three
views (BL/B/BR). The same backbone produces depth, semantic, and
instance predictions selected by textual task prompt. View labels follow
nuScenes convention.}
\label{fig:qual_2d_b}
\end{figure}

\begin{figure}
\centering
\setlength{\tabcolsep}{2pt}
\renewcommand{\arraystretch}{0.6}
\begin{tabular}{@{}cc@{}}
\small Semantic Prediction & \small Semantic Ground-Truth \\ 
\includegraphics[width=0.4\linewidth]{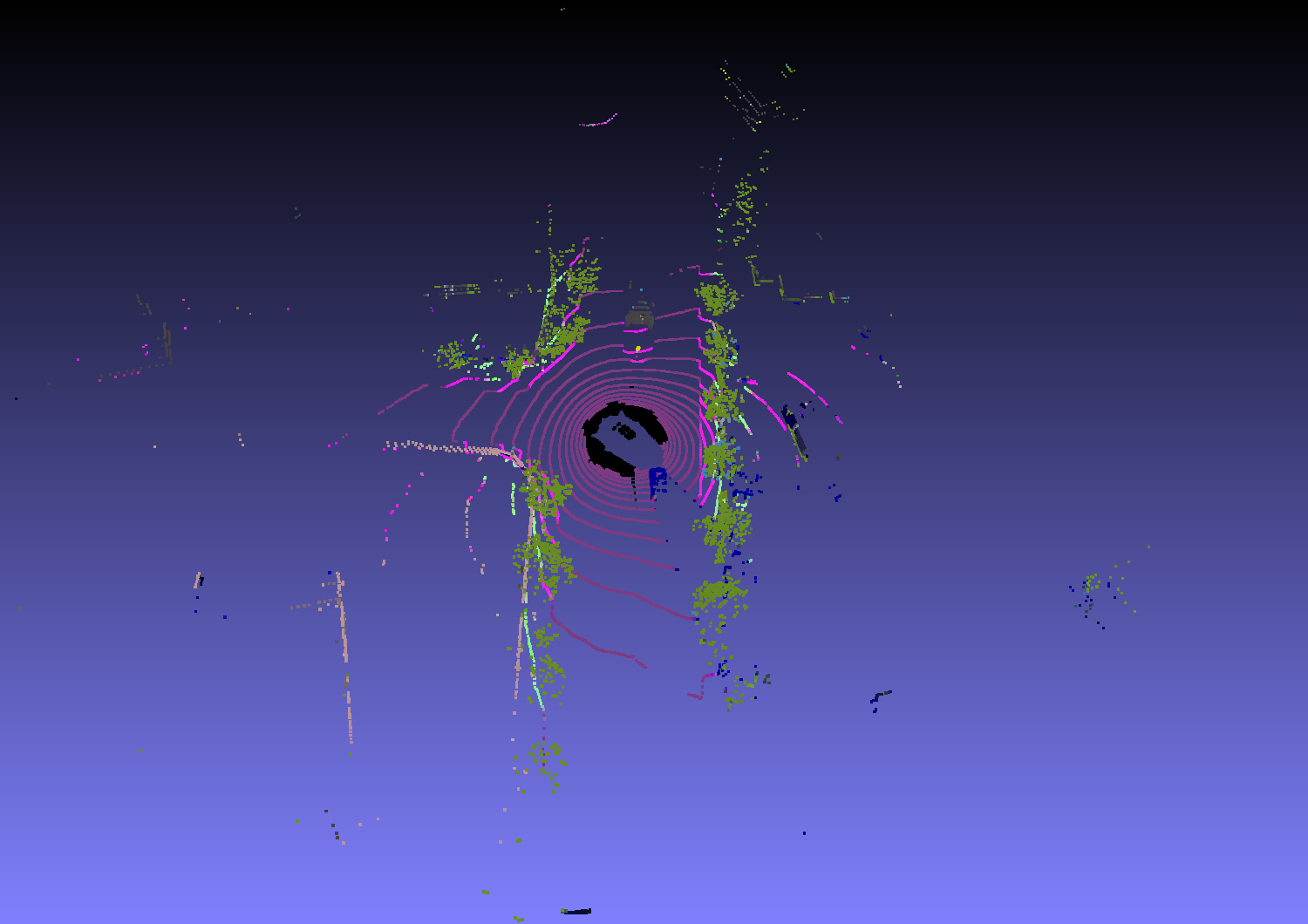} &
\includegraphics[width=0.4\linewidth]{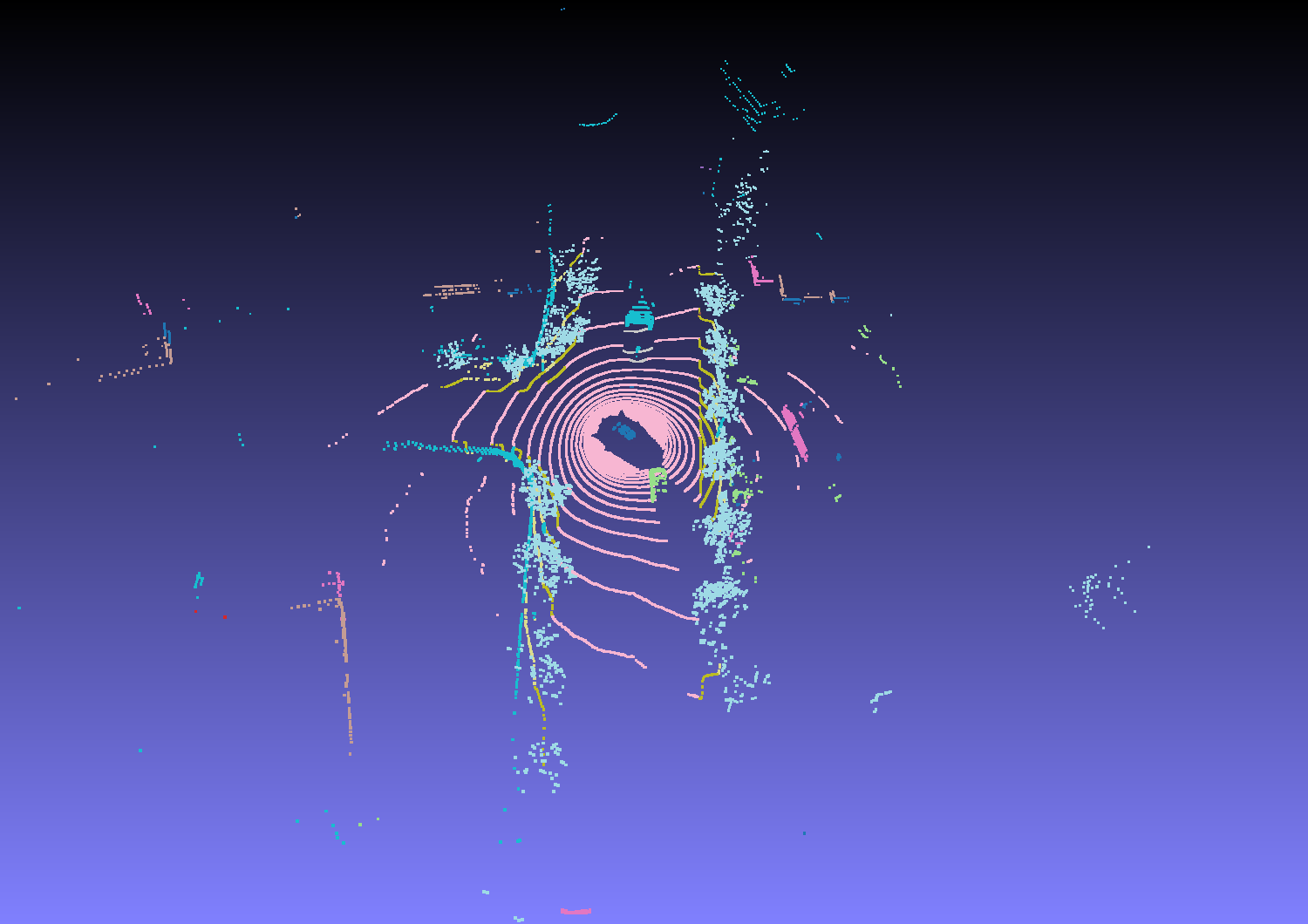} \\
\small Instance Prediction & \small Instance Grount-Truth \\[2pt]
\includegraphics[width=0.4\linewidth]{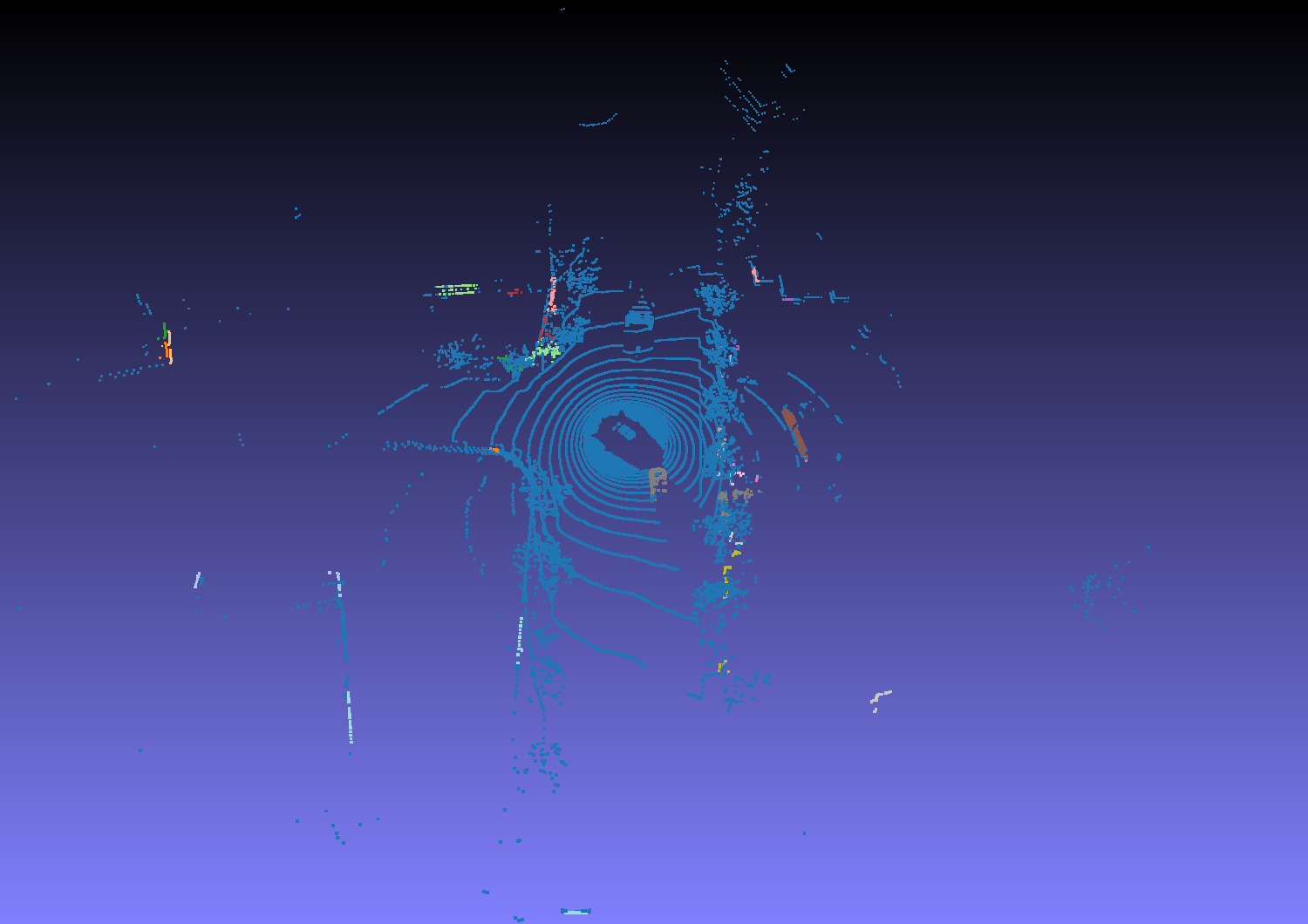} &
\includegraphics[width=0.4\linewidth]{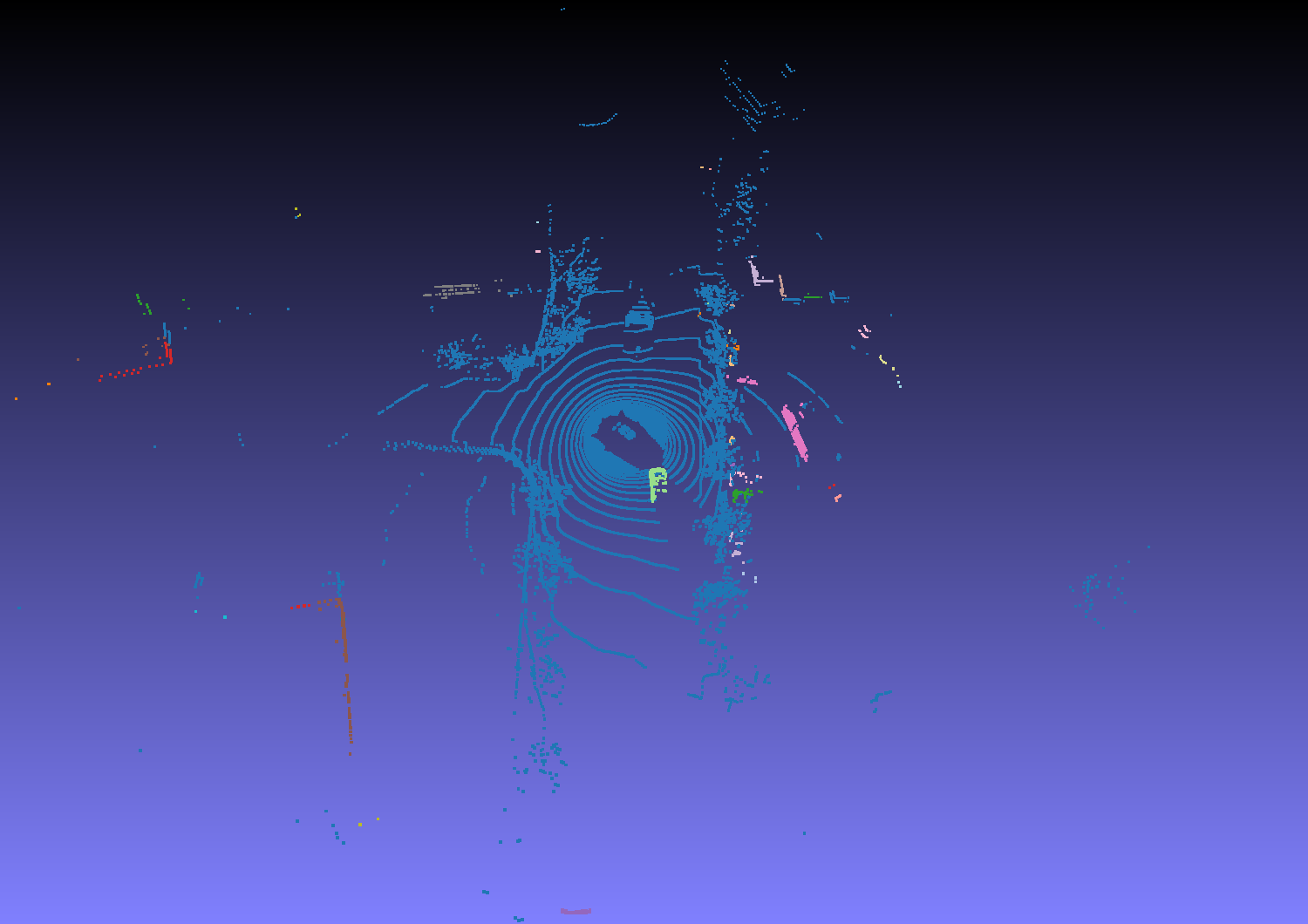} \\

\end{tabular}
\caption{Backprojected 3D outputs vs. ground truth on an additional nuScenes
validation scene. Top row: semantic segmentation and ground truth. Bottom row: instance predictions and ground truth.
Semantic predictions are restricted to the Cityscapes label set. The
ground truth uses a randomly assigned color palette over the nuScenes
lidarseg classes, since the two taxonomies do not share a canonical
color mapping. Instance colors are arbitrary and not matched between
prediction and ground truth, so correctness corresponds to the spatial
separation of distinct objects rather than to color agreement.}
\label{fig:qual_3d}
\end{figure}

\begin{figure}
\centering
\setlength{\tabcolsep}{2pt}
\renewcommand{\arraystretch}{0.6}
\begin{tabular}{@{}cc@{}}
\small Sparse LiDAR input & \small Densified output \\[2pt]
\includegraphics[width=0.4\linewidth]{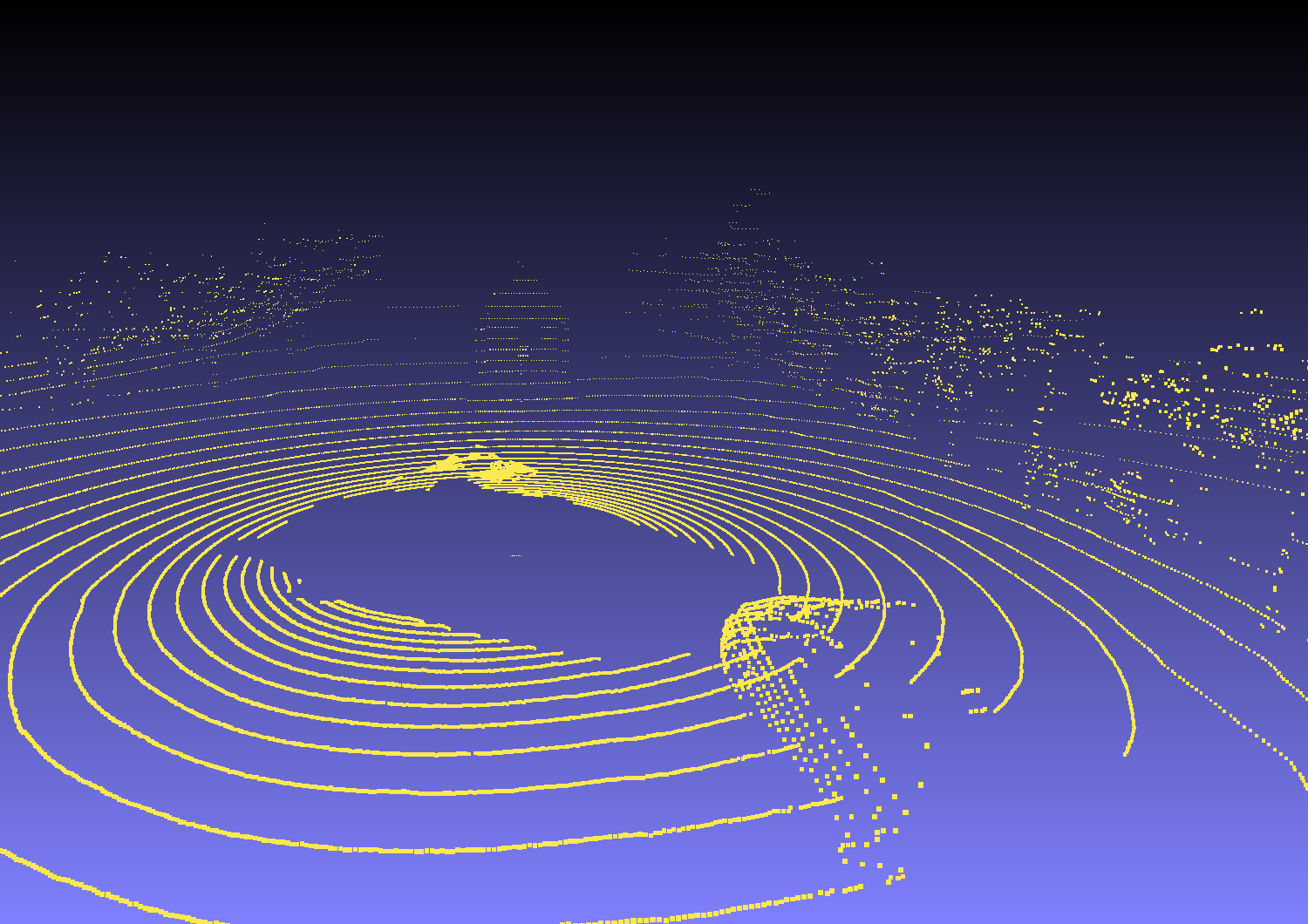} &
\includegraphics[width=0.4\linewidth]{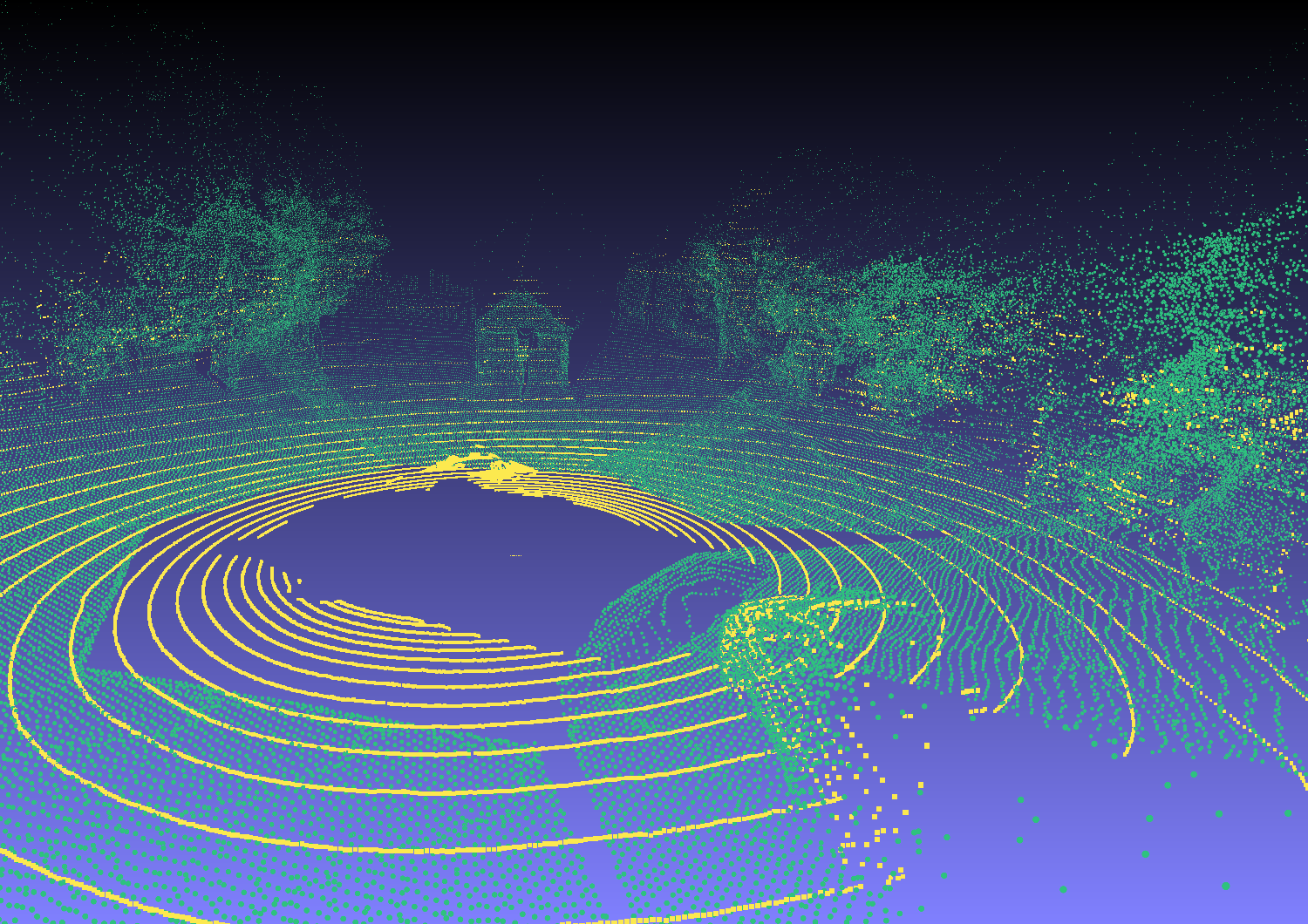} \\
\end{tabular}
\caption{Depth densification quality. Left: original sparse LiDAR scan.
Right: dense depth produced by the model and aligned to metric scale
via least-squares regression against the input LiDAR depth
(Section~\ref{sec:depth}).}
\label{fig:qual_depth_dense}
\end{figure}

We provide a complete qualitative output for a further scene on the validation
set. Figure~\ref{fig:qual_2d_b} shows the model's direct 2D outputs
across all six camera views. Figure~\ref{fig:qual_3d} shows the
backprojected 3D outputs against ground truth.
Figure~\ref{fig:qualitative_pca} shows feature \glspl{pca}.
Figure~\ref{fig:qual_depth_dense} shows the densified depth output after
least-squares metric alignment.
\begin{figure}
\centering
\setlength{\tabcolsep}{2pt}
\renewcommand{\arraystretch}{0.6}
\begin{tabular}{@{}ccc@{}}
\small Depth prompt & \small Semantic prompt & \small Instance prompt \\[2pt]
\includegraphics[width=0.33\linewidth]{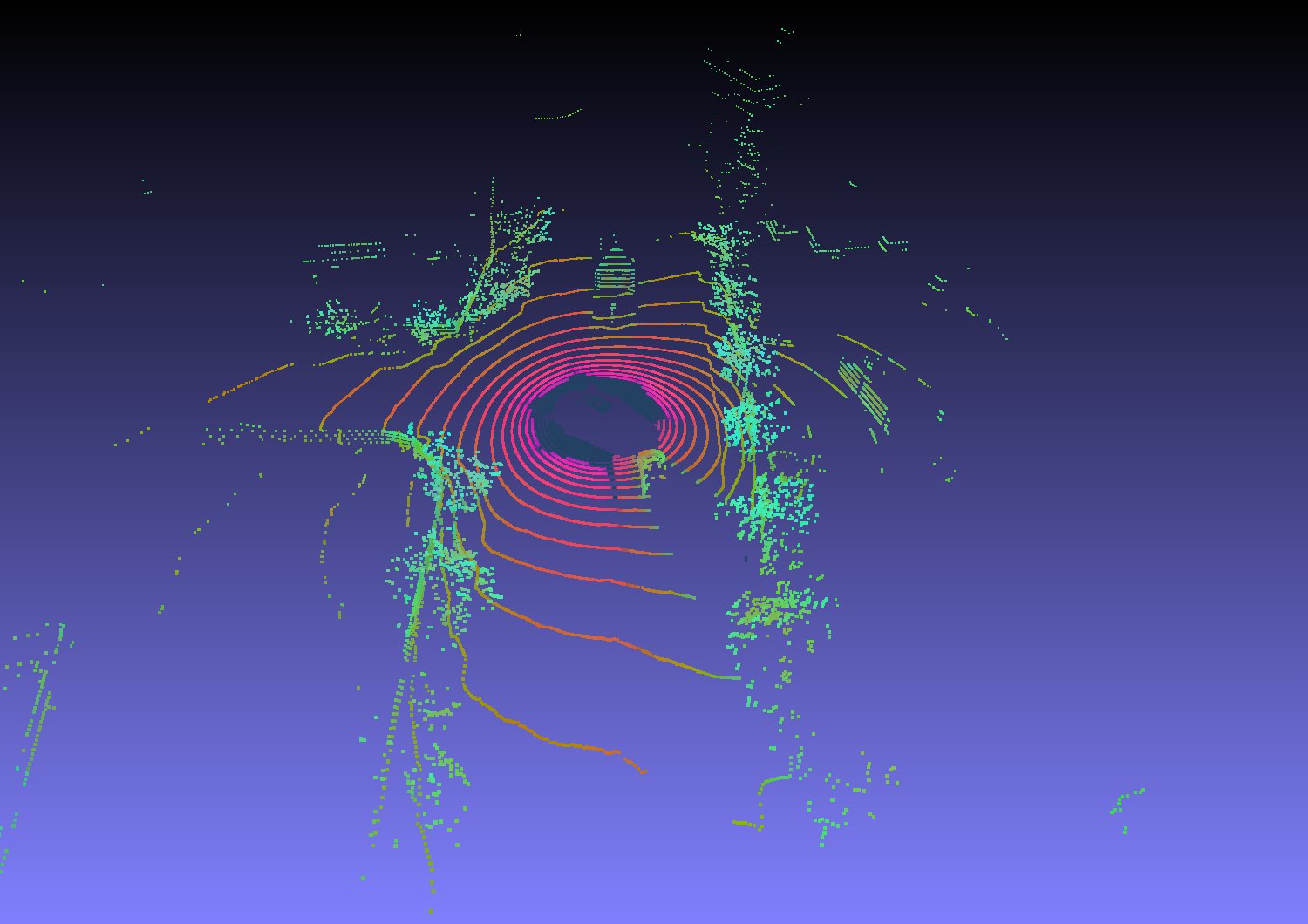} &
\includegraphics[width=0.33\linewidth]{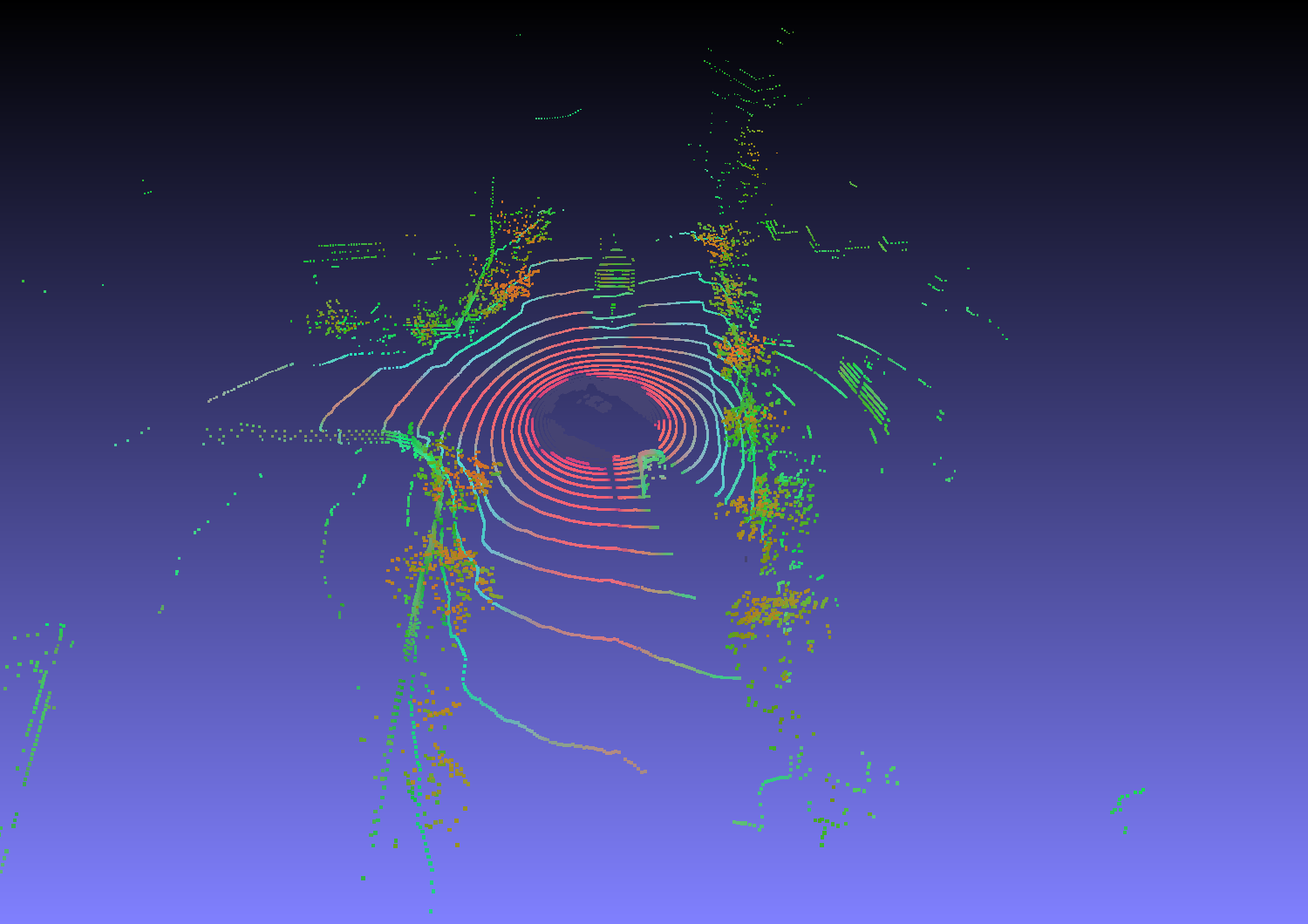} &
\includegraphics[width=0.33\linewidth]{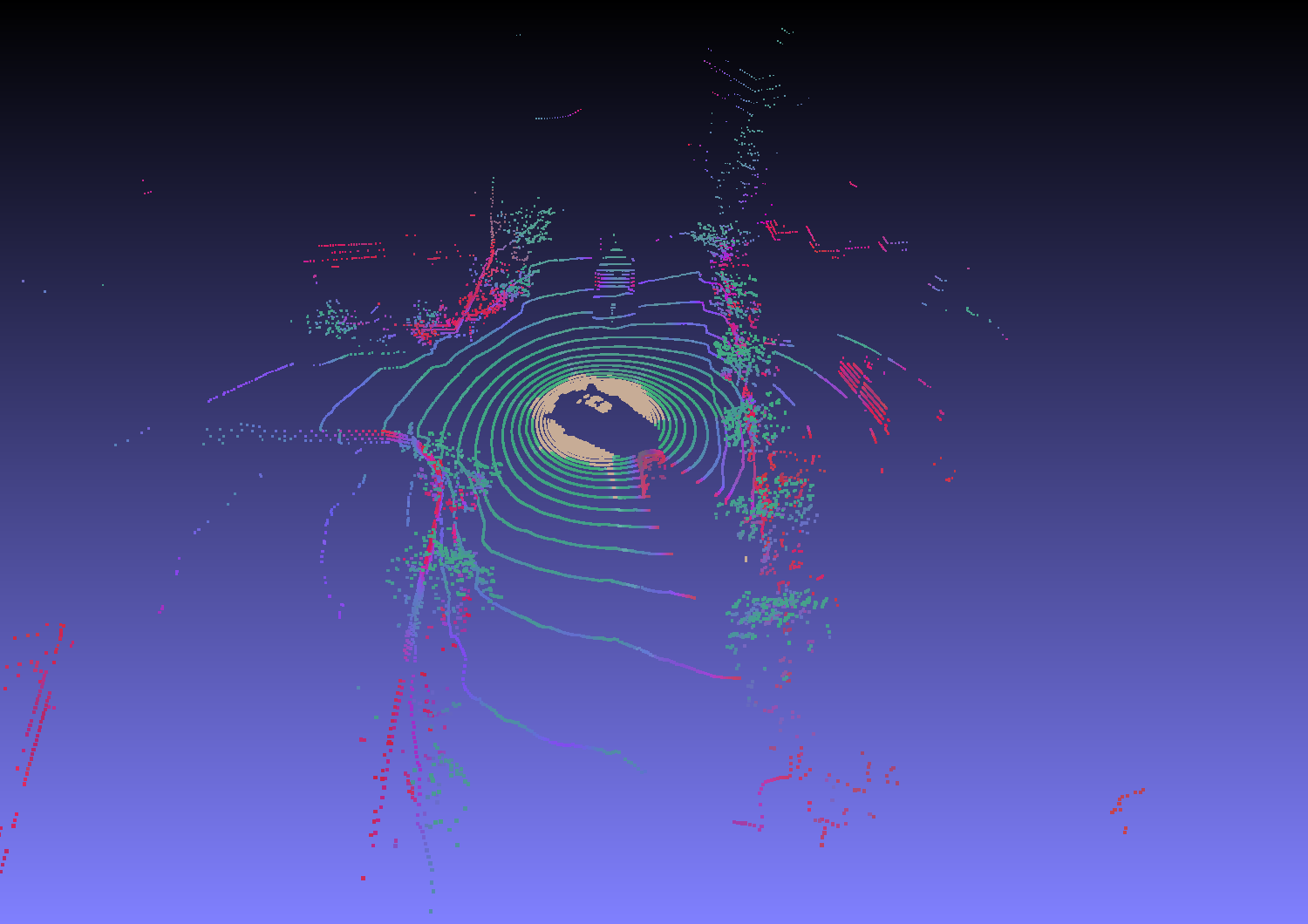} \\
\end{tabular}
\caption{PCA visualization of decoder level 2 features under each task
prompt, mapped to RGB on the backprojected LiDAR point cloud. Same
prompt and level as the main-text Figure~\ref{fig:qualitative_pca}.}
\label{fig:qual_pca_extended}
\end{figure}

\end{document}